\documentclass[10pt,twocolumn,letterpaper]{article}

\usepackage[pagenumbers]{cvpr} 

\usepackage{graphicx}
\usepackage{amsmath}
\usepackage{amsthm}
\usepackage{amssymb}
\usepackage{booktabs}
\usepackage{arydshln}
\usepackage{multirow}
\usepackage{algorithm}
\usepackage{algorithmic}

\usepackage{color}
\usepackage{pifont}

\newcommand{\dmem}{D^\text{mem}}
\newcommand{\soft}{\operatorname{softmax}}
\newcommand{\rld}{\mathcal{L}_{\text{RLD}}}
\newcommand{\lkd}{\mathcal{L}_{\text{kd}}}
\newcommand{\lce}{\mathcal{L}_{\text{ce}}}

\newcommand{\smaller}[1]{\fontsize{7pt}{1em}\selectfont{#1}}
\newcommand{\tablestyle}[2]{\setlength{\tabcolsep}{#1}\renewcommand{\arraystretch}{#2}\centering\footnotesize}

\newlength\savewidth\newcommand\shline{\noalign{\global\savewidth\arrayrulewidth
  \global\arrayrulewidth 1pt}\hline\noalign{\global\arrayrulewidth\savewidth}}
  
\usepackage[pagebackref,breaklinks,colorlinks]{hyperref}

\usepackage[capitalize]{cleveref}
\crefname{section}{Sec.}{Secs.}
\Crefname{section}{Section}{Sections}
\Crefname{table}{Table}{Tables}
\crefname{table}{Tab.}{Tabs.}

\def\cvprPaperID{1943} 
\def\confName{CVPR}
\def\confYear{2022}

\begin{document}
\title{Revisiting Catastrophic Forgetting in Class Incremental Learning}

\author{Zixuan Ni\thanks{These authors contributed equally.}\\
Zhejiang University\\
{\tt\small zixuan2i@zju.edu.cn}
\and
Haizhou Shi$^*$\\
Zhejiang University\\
{\tt\small shihaizhou@zju.edu.cn}
\and
Siliang Tang\thanks{Corresponding author}\\
Zhejiang University\\
{\tt\small siliang@zju.edu.cn}
\and
Longhui Wei\\
Huawei Cloud\\
{\tt\small longhuiwei@pku.edu.cn}
\and
Qi Tian\\
Huawei Cloud\\
{\tt\small tian.qi1@huawei.com }
\and
Yueting Zhuang\\
Zhejiang University\\
{\tt\small yzhuang@zju.edu.cn}
}
\maketitle

\begin{abstract}
Although the concept of catastrophic forgetting is straightforward, there is a lack of study on its causes. In this paper, we systematically explore and reveal three causes for catastrophic forgetting in Class Incremental Learning~(CIL). From the perspective of representation learning, \textbf{(i) intra-phase forgetting} happens when the learner fails to correctly align the same-phase data as training proceeds and \textbf{(ii) inter-phase confusion} happens when the learner confuses the current-phase data with the previous. From the task-specific point of view, the CIL model suffers from the problem of \textbf{(iii) classifier deviation}. We observe that there is only a few works on how to alleviate the inter-phase confusion. To initiate the research on this specific issue, we propose a simple yet effective framework, \textbf{C}ontrastive \textbf{C}lass \textbf{C}oncentration for \textbf{CIL}~(C4IL). 
Our framework leverages the class concentration effect of contrastive learning, yielding a representation distribution with better intra-class compactibility and inter-class separability. Empirically, we observe that C4IL significantly lowers the probability of inter-phase confusion and as a result improves the performance on multiple CIL settings of multiple datasets.

\end{abstract}


\section{Introduction}
\label{sec:intro}

Unlike the traditional deep learning paradigm that trains the model on the full dataset, in Class Incremental Learning (CIL), the model is continually trained on the new-class data, during which process the old-class data is unavailable~\cite{rebuffi2017icarl,li2017learning,parisi2019continual,zhao2020maintaining,liu2020mnemonics,zhang2020class,castro2018end}. Directly fine-tuning the model on the newly added data will result in a sharp decline of the classification accuracy on the old classes~\cite{li2017learning}. This phenomenon is called catastrophic~forgetting~\cite{goodfellow2013empirical}. Although this concept is easy to understand, only a few works discuss its complex causes~\cite{hou2018lifelong,lesort2019regularization,mittal2021essentials}. These works either have mere conjectures and lack experimental verification~\cite{hou2018lifelong,lesort2019regularization}, or mention the concept of intra-task/inter-task learning and fails to provide a systematic analysis~\cite{mittal2021essentials}. In summary, the shared understanding of how the existing methods approach catastrophic forgetting has not been reached. 


\begin{figure}
	\centering
	\includegraphics[width=3.2in]{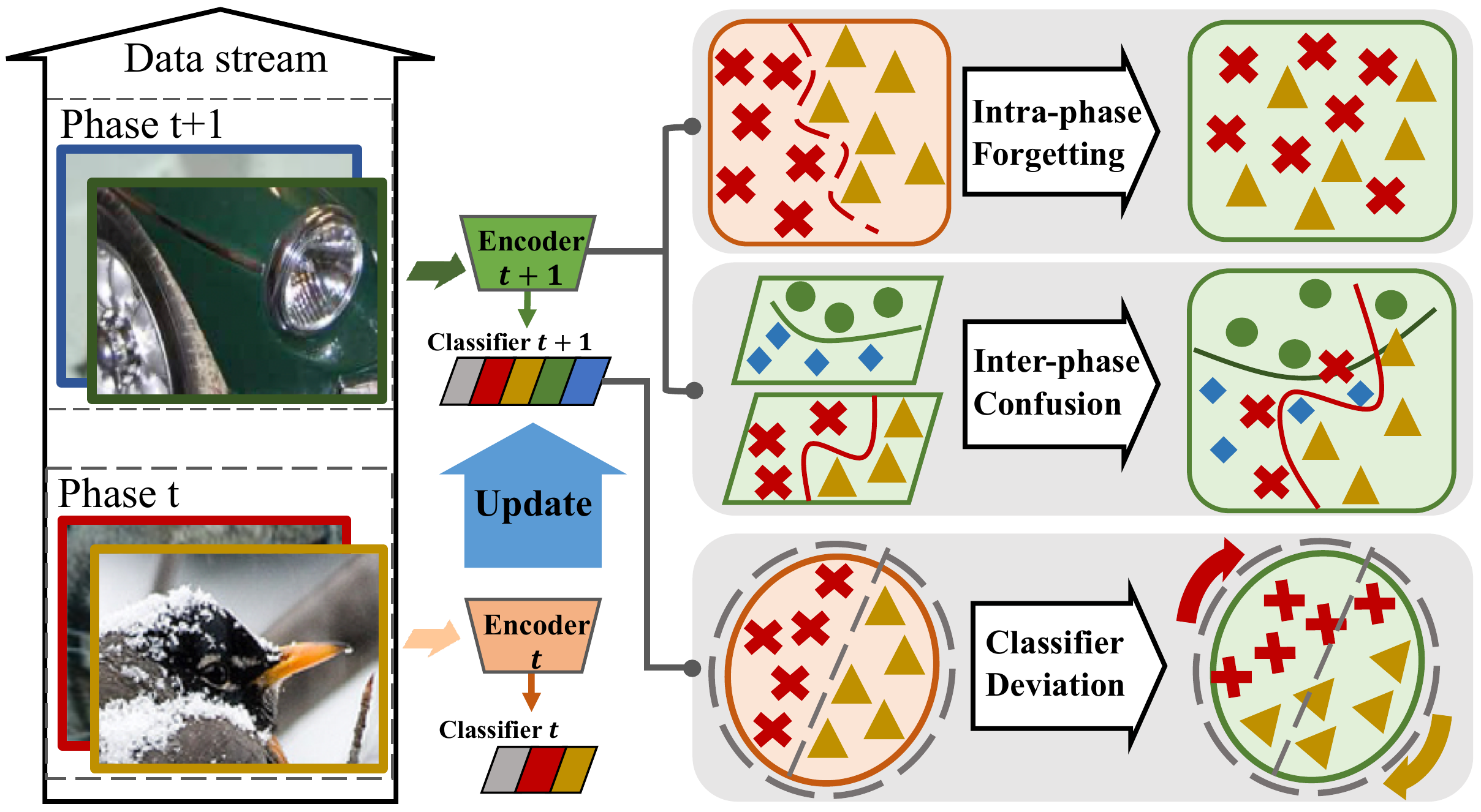}
	\caption{Three causes for catastrophic forgetting in CIL. \textbf{Intra-phase Forgetting}: the model fails to preserve the linear separability of the old tasks' representation spaces. \textbf{Inter-phase Confusion}: the model correctly aligns the data within the same phase, but confuses the data belonging to different phases. \textbf{Classifier Deviation}: the mismatch between the representation space and the classifier gradually happens as CIL proceeds.}
	\label{three_reason}
\end{figure}

The first part of this paper seeks to answer the above two questions: (i)~what comprises the general concept of catastrophic forgetting? and (ii)~what specific aspects do the existing techniques address? 
In CIL, the network parameters before the penultimate layer are shared across different phases, while the classifier's update is class-specific: learning a sample of class $A$ will not affect the class $B$'s classifier. Therefore it is natural to view the catastrophic forgetting on the representation level and the downstream-task level separately. On the representation level, we introduce the concept of the linear separability to help us analyze the variation of the representation space during CIL. We then disentangle the representation-level forgetting into two main causes: \textbf{(i)~intra-phase forgetting} and \textbf{(ii)~inter-phase confusion}.
As shown in Fig.\ref{three_reason}, intra-phase forgetting refers to the phenomenon that the model fails to preserve the quality of the old tasks' representation spaces as the training proceeds; inter-phase confusion refers to the phenomenon that the model mistakenly aligns different tasks representation spaces together. On the downstream-task level, \textbf{(iii)~classifier deviation} refers to the fact that when the model is trained on the new task, the representation space is rotated and shifted, during which phase the old classifier is not updated, and thus there is a mismatch between the representation space and the classifier. 

The existing strategies address the three causes of forgetting in different ways. By a series of experiments, we find that the distillation technique proposed by the classic LwF \cite{li2017learning} addresses the problem of intra-phase forgetting. Meanwhile, multiple methods~\cite{castro2018end,yu2020semantic,zhao2020maintaining} explicitly address the problem of classifier deviation and yield a more up-to-date classifier during CIL training. On the other hand, the problem of inter-phase confusion is far less studied and is only explicitly addressed in limited scenarios~\cite{lesort2019regularization}.
To initiate the research on approaching this specific issue, we propose a simple yet effective framework, \textbf{C}ontrastive \textbf{C}lass-\textbf{C}oncentration for \textbf{CIL}~(C4IL). It leverages the label-guided contrastive learning as a measure to form a more compact representation distribution, so that the probability of confusing the data of different phases is lowered. 
Empirical evaluation shows that C4IL, targeting specifically at the problem of inter-phase confusion, can significantly alleviate catastrophic forgetting for the CIL model and outperforms the existing methods. The improvement brought by C4IL is also flexible: the framework can be plugged in both memory-free (NoMem) and memory-based (Mem) methods and achieve better results. Qualitative evaluation also demonstrates that our method produces a more compact representation distribution that alleviates the confusion problem.

In summary, the main contributions of this work are:
\begin{itemize}
    \item We introduce the measure of representation space and decouple the complex concept of catastrophic forgetting into three individual causes. 
    \item We identify how the existing techniques address these three causes and point out the importance of addressing inter-phase confusion by empirical experiments.
    \item We propose a simple yet effective CIL method to explicitly prevent inter-phase confusion by improving the model's concentration ability.
\end{itemize}

\section{Catastrophic Forgetting in CIL}
\noindent\textbf{Related work.}\quad
The most popular methods in CIL are either based on regularization strategy~\cite{li2017learning,rebuffi2017icarl,hu2021distilling,zhang2020class,douillard2020podnet,yu2020semantic,kirkpatrick2017overcoming} or the rehearsal strategy \cite{zhao2020memory,rebuffi2017icarl,castro2018end,liu2020mnemonics,douillard2020podnet,pomponi2020pseudo}, or the combination of the two \cite{hou2019learning,zhao2020memory,liu2020more,zhao2020maintaining,he2020incremental}. The regularization strategy adopts the previous-phase model(s) as a constraint for the current-phase training, which helps the model to preserve the old knowledge. LwF \cite{li2017learning} first introduces the knowledge distillation~\cite{hinton2015distilling} to continual learning and yields great improvement. Ever since then this technique become default and are adopted by many works~\cite{rebuffi2017icarl,hou2019learning,hu2021distilling,douillard2020podnet,hou2019learning,zhao2020memory,liu2020more,zhao2020maintaining,he2020incremental}. The rehearsal strategy, on the other hand, directly stores a small amount of previous-phase data in the memory bank and replay them during training. By doing so, it converts the problem of catastrophic forgetting into the problem of data imbalance and how to select the most representative samples~\cite{castro2018end,wu2019large,hu2021distilling,wu2019large}, which somewhat weakens the significance of their finding as a measure to address catastrophic forgetting. To avoid this, the pseudo-rehearsal strategy regenerates the previous-phase data with the popular generative models~\cite{he2018exemplar,zhao2021memory}. \\

\noindent\textbf{A representation learning perspective.}\quad
Although the classification accuracy in CIL has been significantly improved, there is still a lack of shared understanding of what comprises the catastrophic forgetting and how the aforementioned methods address it~\cite{hou2018lifelong,benzing2020understanding,kirkpatrick2017overcoming}. In this section, we will take the viewpoint of representation learning and introduce the metric of linear evaluation accuracy to measure the representation quality of the penultimate layer of the CIL models~\cite{zhang2016colorful, chen2020simple, he2020momentum, oord2018representation, tian2019contrastive}. The basic assumption behind linear evaluation is that the quality of a representation space has a positive correlation with its linear separability. Specifically, we retrain only the linear classifier of a model, and the accuracy yielded is the indicator of its representation space quality.

With the help of linear evaluation, we will firstly decouple the catastrophic forgetting into representation-level forgetting, which consists of intra-phase forgetting and inter-phase forgetting, and downstream-level forgetting, namely classifier deviation. We will then analyze how the prevalent strategies address them by empirical studies.

\subsection{Intra-phase Forgetting}
\begin{figure}[h]
\vspace{-1em}
\centering
\includegraphics[width=2.5in]{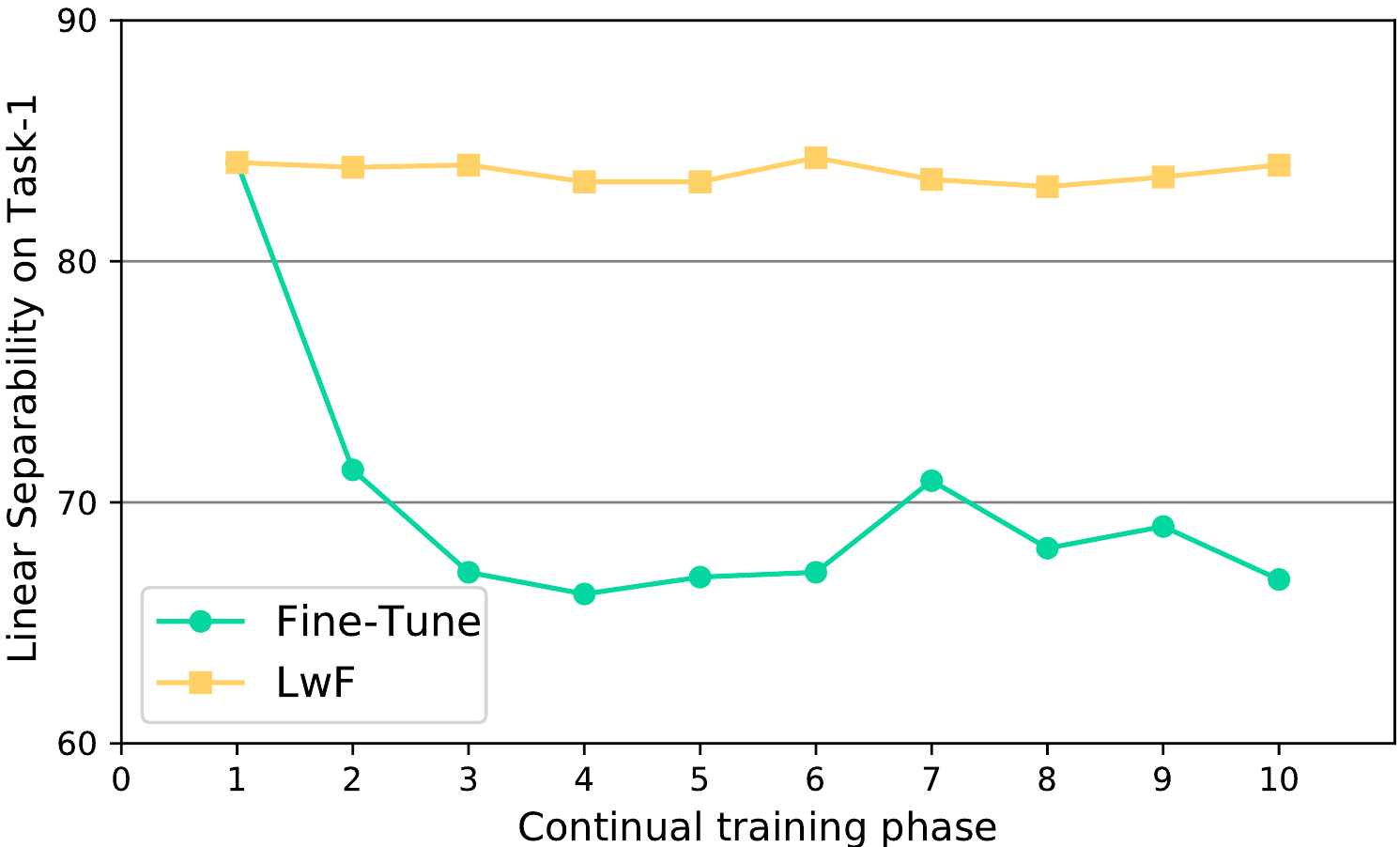}
\vspace{-0.5em}
\caption{Linear evaluation accuracy of the phase-1 data during CIL, evaluated on CIFAR-100~\cite{krizhevsky2009learning}. Distillation method LwF~\cite{li2017learning} perfect addresses the problem of intra-phase forgetting.}
\label{inter_phase_forgetting}
\end{figure}
Intra-phase forgetting denotes how much the model forgets about mapping the data of the same phases into the representation space as the CIL training proceeds. Without loss of rigor, we evaluate the phase-1 data linear separability of a vanilla fine-tuning method~(which is trained on a series of tasks sequentially without any additional measures) and a classic distillation-based method LwF~\cite{li2017learning} on the 10-phase CIL on CIFAR-100~\cite{krizhevsky2009learning}, as shown in Fig.\ref{inter_phase_forgetting}. We find that the vanilla fine-tuning method's linear separability on the phase-1 data distribution drops by a large margin during the training of the CIL~(84.1\%\ding{213}66.8\%). In contrast, the simple LwF method helps to maintain the linear separability of the previous phases~(84.1\%\ding{213}84.3\%). \textbf{Therefore here we conclude that the distillation strategy solves the problem of intra-phase forgetting.} We conjecture that the vanilla fine-tuning method forgets the previous classification task-specific information during CIL and only preserves the general representative power shared by different phases. On the other hand, LwF forces the model to preserve the task-specific information by distillation on the classification distribution, which yields better linear separability. 

However, one must note here, our validation on the efficacy of the distillation-based method in CIL may depend on certain conditions. For example, the current benchmark of the CIL is typically on the same-domain data. And the distillation method's effect on solving the intra-phase forgetting might be undermined when there is an apparent domain shift among different phases~\cite{li2017learning,mittal2021essentials}. This issue is intriguing but beyond the scope of this paper, thus we leave it for future works.

\subsection{Inter-phase Confusion}
\begin{figure}[h]
\centering
\vspace{-1em}
\includegraphics[width=2.5in]{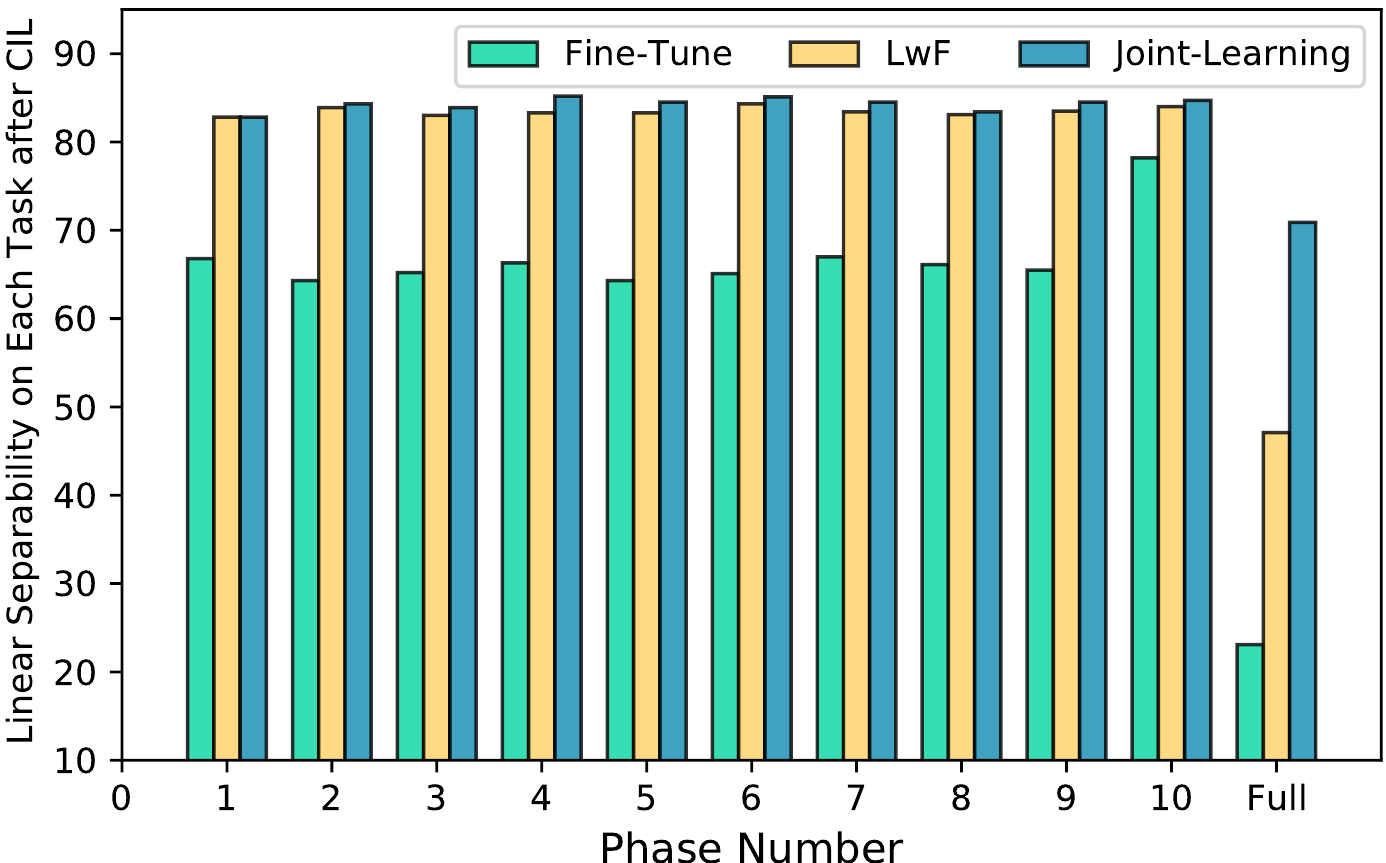}
\vspace{-0.5em}
\caption{Linear separability of each individual phase \textbf{after CIL/joint training}, evaluated on CIFAR-100~\cite{krizhevsky2009learning}. LwF and joint learning's performances are close on individual phases, while LwF is much worse than joint learning when evaluated on the full dataset. It reflects the severe problem of inter-phase confusion.}
\label{intra_phase_confusion}
\end{figure}
Different from intra-phase forgetting, which indicates to what extent the model confuses representation distribution of the same-phase data, inter-phase confusion refers to the phenomenon that the model fails to align different phases' data into different positions due to the constraint of continual learning: the disallowance of preserving the old tasks' data. 

In order to better showcase and study the phenomenon of inter-phase confusion, we evaluate the linear separability of fine-tuning method and LwF after the complete CIL training. We compare its performance with the upper-bound joint-learning method~(in other words, supervised learning). As shown in the Fig \ref{intra_phase_confusion}, when evaluated on individual tasks, LwF has a similar representation quality as joint learning. However, when evaluated on the full dataset, the LwF's decrease of the accuracy is significantly larger than joint learning, which is exactly caused by the problem of inter-phase confusion.

The problem of inter-phase confusion is the core and the most challenging aspect of catastrophic forgetting in CIL, whereas has not received enough attention of the community. The lack of attention on this issue causes the performance bottleneck in CIL. To validate its significance, Sec.\ref{sec:method} will explicitly target on the issue of inter-phase confusion by adopting simple techniques. Despite of the simplicity, our measure manages to alleviate inter-phase confusion and thus achieves better performance. 

\subsection{Classifier Deviation}
\begin{figure}[!h]
\centering
\vspace{-1em}
\includegraphics[width=2.5in]{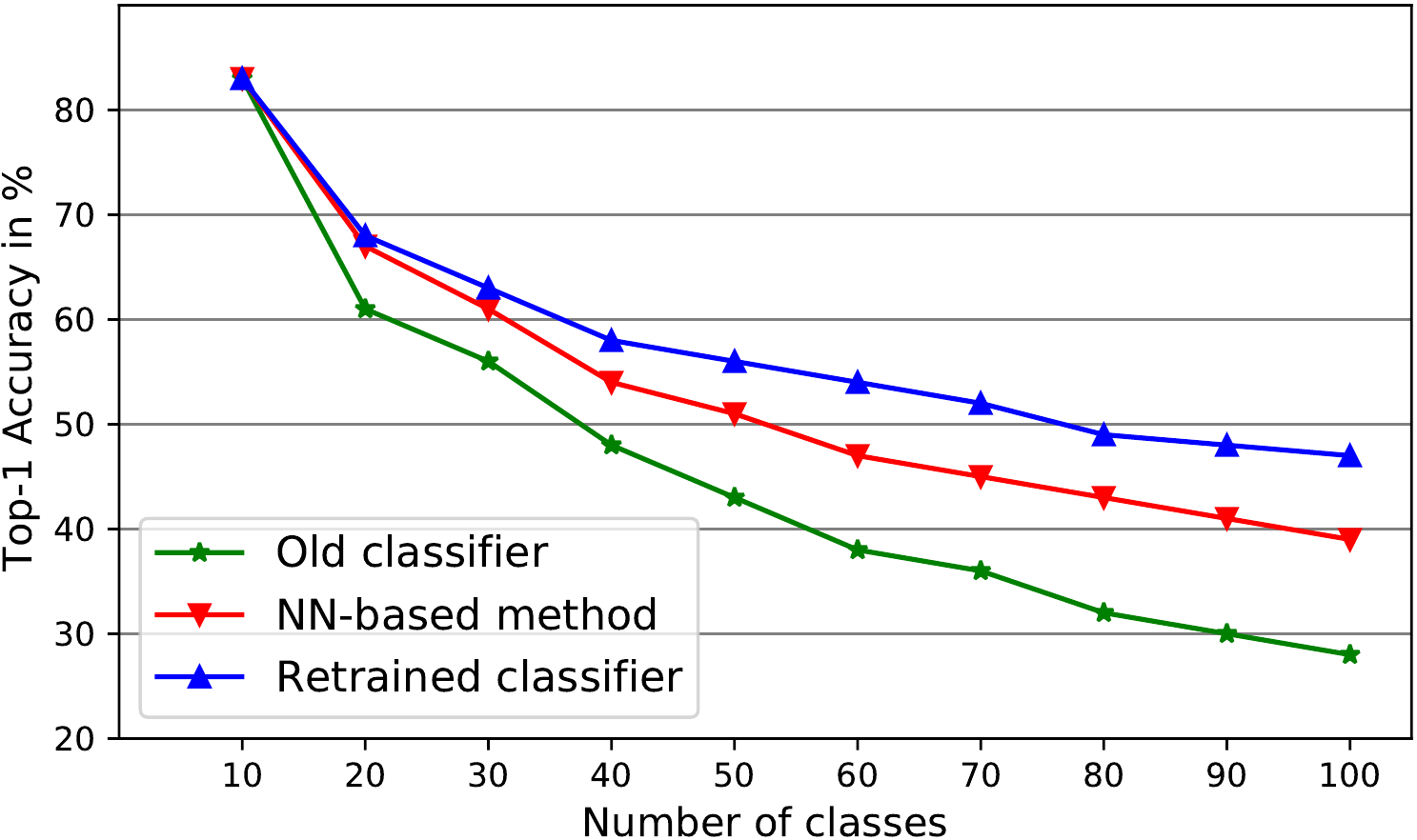}
\vspace{-0.5em}
\caption{Classifier deviation causes the gap between the model's classification accuracy and its linear evaluation accuracy. The NN-based classification methods~\cite{rebuffi2017icarl,zhao2020maintaining,he2020incremental} preserve a small amount of the data to alleviate the problem.}
\label{classifier_deviation}
\end{figure}
There is a significant gap between the classification accuracy (using the old classifier) of the CIL model and its linear evaluation accuracy (using the retrained classifier), as we see in Fig.\ref{classifier_deviation}. In order to alleviate the classifier deviation~\cite{hou2018lifelong}, some works try to estimate the shift of feature space and compensate for it~\cite{yu2020semantic}, whereas most of the current works \cite{rebuffi2017icarl,hou2019learning,hu2021distilling,douillard2020podnet,zhao2020memory,liu2020more,zhao2020maintaining,he2020incremental} preserve a small amount of past data in the memory bank and use NN clustering method NME \cite{rebuffi2017icarl} as a substitution of the classification head.

\begin{figure*}[t]
\begin{subfigure}[c]{0.57\linewidth}
\centering
\includegraphics[width=3.3in]{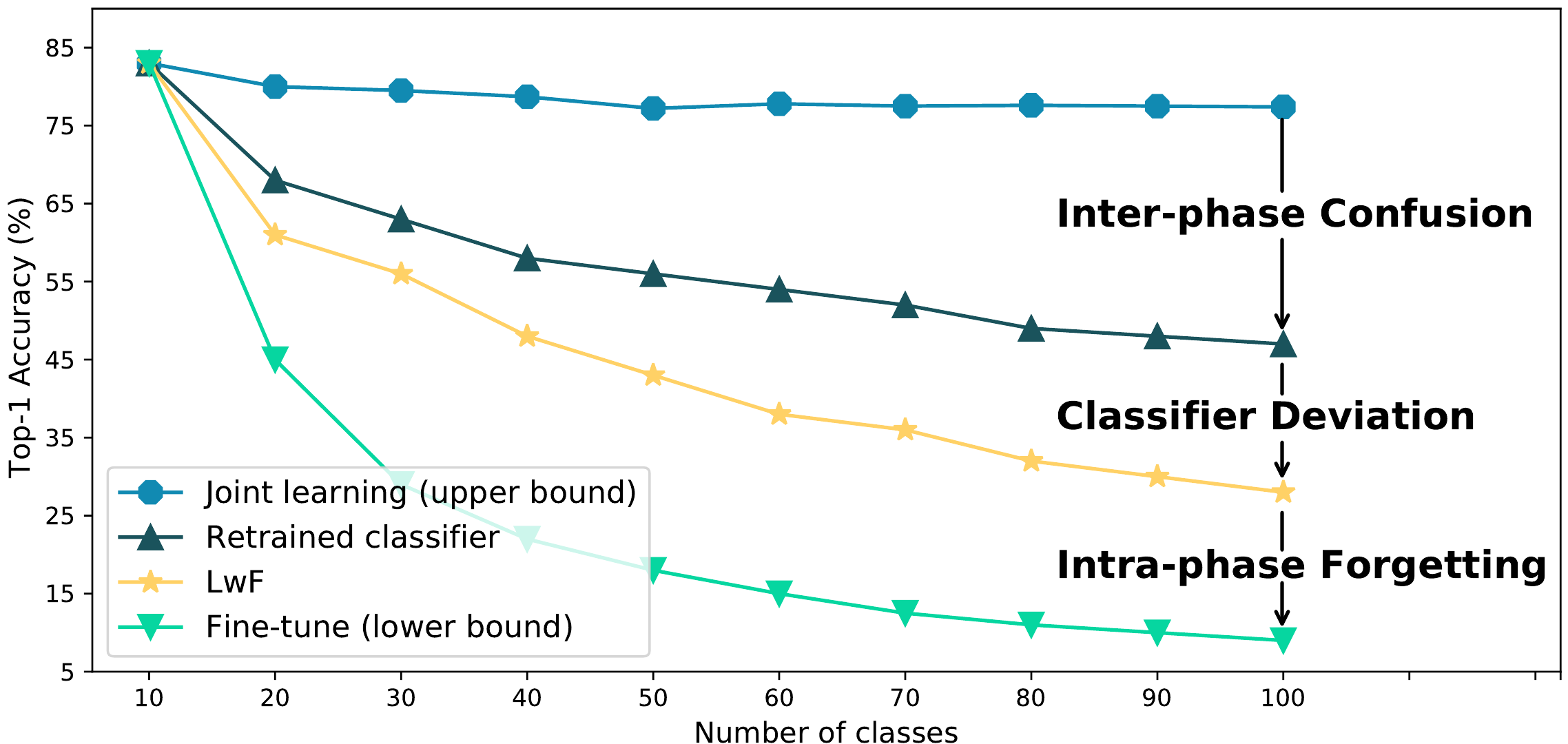}
\label{figure:123}
\end{subfigure}
\begin{subfigure}[c]{0.4\linewidth}
\centering
\small
\tablestyle{3pt}{1.4}
\begin{tabular}{cccc}
\shline
 Methods & \shortstack{Intra-phase \\ Forgetting}  & \shortstack{Inter-phase \\ Confusion} & \shortstack{Classifier \\ Deviation} \\
\hline
{LWF~\cite{li2017learning}}	 &	$\surd$	&	  &   \\ 
{iCaRL~\cite{rebuffi2017icarl}} &	$\surd$	&	 & $\surd$ \\
{BiC~\cite{wu2019large}}  &	$\surd$	 &	 & $\surd$  \\
{ETE~\cite{castro2018end}}  &	$\surd$	 &	 & $\surd$  \\ 
{LUCIR~\cite{hou2018lifelong}} &  $\surd$ &	$\surd$ & $\surd$  \\
\hline
C4IL(ours) &	$\surd$	 &	$\surd$  & $\surd$  \\
\shline
\end{tabular}
\vspace{1.2em}
\label{tab:123}
\end{subfigure}
\caption{Composition of the general catastrophic forgetting in CIL and how the current measures address them. 
\textbf{Left figure:} catastrophic forgetting is decomposed to three causes. The (i)~intra-phase forgetting is solved by distillation technique as in LwF~\cite{li2017learning}. By retraining a linear classifier on top of the encoder network, we exempt the influence of (iii)~classifier deviation from our analysis. The remaining gap between the blue line and the purple line~(joint learning upper-bound) is purely caused by (ii)~inter-phase confusion. 
\textbf{Right table:} aspects of ctastrophic forgetting the existing methods explicitly address. We do not account the memory bank as an explicit measure.}
\label{fig:summary}
\end{figure*}

\subsection{Summary}
In this section, we have identified the key factors that comprise catastrophic forgetting. As shown in Fig.\ref{fig:summary}, the general concept of catastrophic forgetting in CIL is strictly composed of (i)~intra-phase forgetting, (ii)~inter-phase confusion, and (iii)~classifier deviation. By taking the distillation technique as in LwF~\cite{li2017learning}, we solve the intra-phase forgetting and improve the vanilla fine-tuning method by 18.7 points of accuracy~(9.5\ding{213}28.2). Then we measure the accuracy a LwF model can achieve with absolutely no classifier deviation by retraining the classifier on the full data. It yields a improvement of 19.0 points~(28.2\ding{213}47.2). The final performance gap of 30.2 points~(47.2\ding{213}77.4) between this method and the upper-bound joint learning can be concluded as the consequence of inter-phase confusion, indicating a huge potential in explicitly addressing this issue in CIL. 

In the following sections, as a direct evidence of the importance of inter-phase confusion, we propose a simple yet effective framework, Contrastive Class Concentration for CIL~(C4IL): we leverage the class concentration effect of contrastive learning to yield a representation distribution of more intra-class compactbility and inter-class separability, thereby reducing the probability of inter-phase confusion in CIL. C4IL's components work on the representation level, making it an easy-to-plugin module that can be directly applied to both memory-free and memory-based frameworks.


\section{Alleviate Inter-phase Confusion in CIL}
\label{sec:method}
\subsection{Problem Definition}
\noindent\textbf{General CIL setting.}\quad Suppose that the dataset $D$ constains training data $X$ and labels $Y$. $C$ is the set containing all labels. 
We split $D$ into $N$ sub-datasets $\{D^{(1)},...,D^{(N)}\}$ to simulate a stream of data and $D^{(t)}$ denotes the data in the incremental phase $t$, where $t \in \{ 1,2,3,...,N\} $. The sub-dataset $D^{(t)} =\{ (x^{(t)}_1,y^{(t)}_1),...,(x^{(t)}_n,y^{(t)}_n)|x^{(t)}_j \in X^{(t)},y^{(t)}_j \in C^{(t)} \}$ where $X^{(t)}$ means the training data in $D^{(t)}$ and $C^{(t)}$ means the sub-classes in $C$ and $n^{(t)}$ is the number of data in $D^{(t)}$. Typically for different sub-datasets, $X^{(i)} \cap X^{(j)} = \emptyset $ and $C^{(i)} \cap C^{(j)} = \emptyset $ for $i \neq j$. 
When the model is trained during the incremental phase $t$, the previous sub-datasets $\{D^{(1)},...,D^{(t-1)}\}$ are no longer available. 
Most of the existing methods address a relaxation setting of the CIL: they store a small number of the old data $\dmem$ in the memory bank, and add them into $D^{(t)}$ forming $D^{(t*)}$, where $D^{(t*)}=D^{(t)}\cup \dmem$.\\

\begin{figure*}[t]
	\centering
	\includegraphics[width=1\linewidth,height=2.8in]{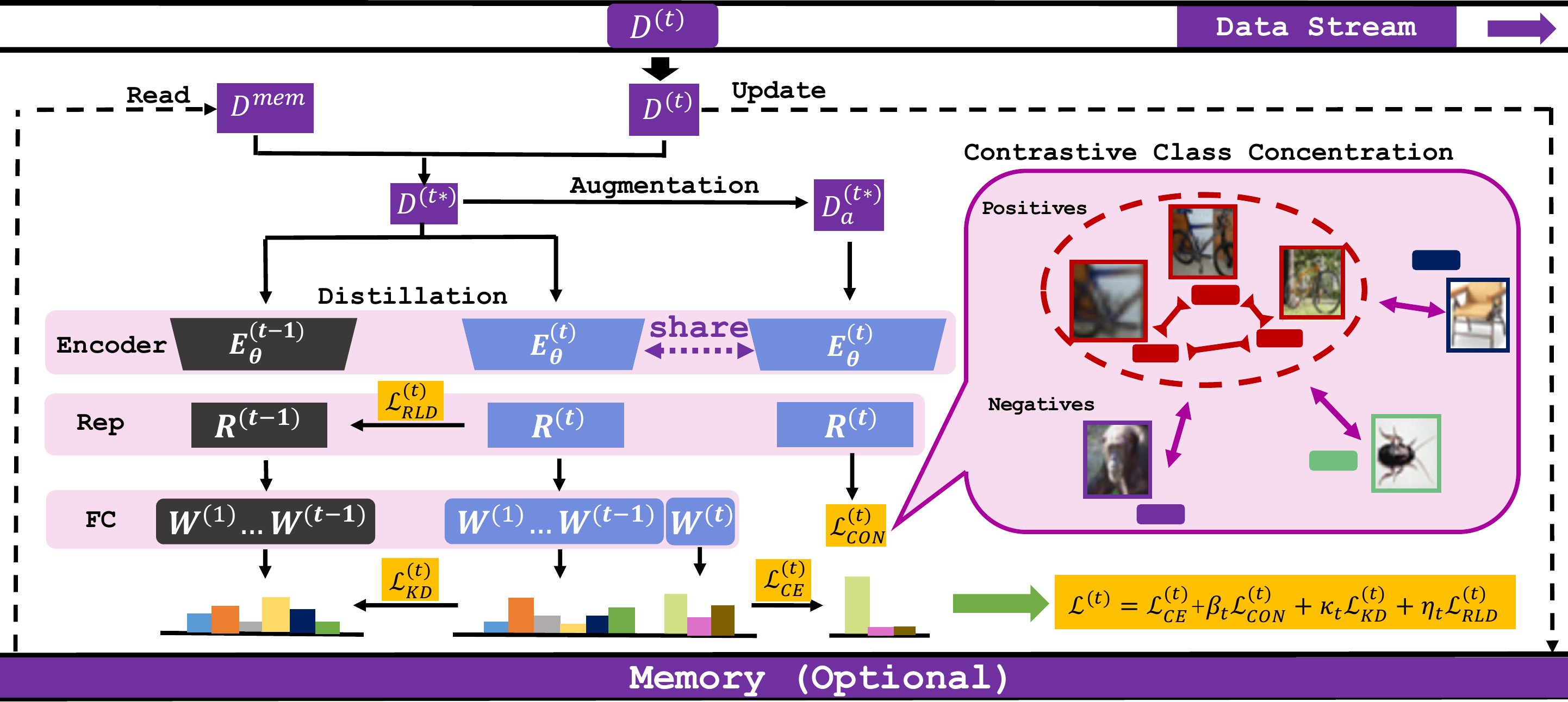}
	\caption{Illustration of the process of C4IL. We use $E_{\theta}^{(t)}$ and $W^{(t)}$ to represent the Encoder and the weights of classifier in phase $t$. $R^{(t)}$ is the representation from $E_{\theta}^{(t)}$. Firstly, we read the stored data from memory (if required) and yield the dataset $D^{(t*)}$ at phase $t$. Then we train the model with the cross-entropy of the prediction and ground-truth $\lce^{(t)}$, and two-level distillation objectives $\rld^{(t)}$ and $\lkd^{(t)}$. In addition, to achieve better concentration effect, we introduce the contrastive class concentration loss $\mathcal{L}^{(t)}_\text{con}$ to pull together same-class representations and push away dissimilar ones. Finally, after phase $t$ is completed, we update the memory by replacing some of the old samples (randomly) with some of the new ones (if required).} 
	\label{training_process}
	\vspace{-0.3em}
\end{figure*}

\noindent\textbf{Dividing representation and classification.}\quad To model the representation-level and downstream-level forgetting, we divide the model into an encoder network $E_{\theta}^{(t)}$ and a linear classifier $W^{(t)}$ at phase $t$. We denote the representation set generated at phase $t$ by $R^{(t)}=\{r^{(t)}_j\}_{j=1}^{n^{(t)}}$ where $r^{(t)}_j=E_\theta^{(t)}(x^{(t)}_j)$ is the representation of $x^{(t)}_j$ generated by the encoder network. For the sake of convenience, we denote the representation produced by the previous-phase encoder $E_\theta^{(t-1)}$ as $R^{(t-1)}=\{r^{(t-1)}_j\}_{j=1}^{n^{(t)}}$ where $r^{(t-1)}_j=E_\theta^{(t-1)}(x^{(t)}_j)$.
After the representation is produced, we compute its classification probability over the label set $C^{(t)}$ by $\soft\left(\big[W^{(t)}\big]^\top r^{(t)}_j\right)$. Similarly, the probability over the full label set $C$ is computed as $\soft\left(W^\top r^{(t)}_j\right)$ where $W = [W^{(1)}, W^{(2)}, \cdots, W^{(N)}]$ is the concatenation of all the classification heads.

\subsection{Contrastive Class Concentration}\label{section3.2}
In the field of self-supervised learning~\cite{wu2018unsupervised, oord2018representation, tian2019contrastive, chen2020simple, he2020momentum}, the contrastive loss, i.e. InfoNCE~\cite{oord2018representation} loss pull together the representations that are semantically close~(positive pairs) and push apart the representations of different instances~(negative pairs),
\begin{equation}
\mathcal{L} = {E}\left[-\log \frac{\exp(s(x, x^+) )}{\exp(s(x, x^+))+\sum^{K}_{k=1} \exp(s(x, x^-_k))}\right],
\end{equation}
where the positive sample $x^+$ are obtained by applying aggressive augmentation, i.e. a series of spatial transformations and color transformations, \cite{chen2020simple} on the original input $x$, and $K$ stands for the size of the negative samples. Function $s$ measures the similarity between two data samples by the dot product of their $l2$-normalized representations, namely,
\begin{equation}
s(x_i, x_j) = \frac{E_\theta(x_i)^\top E_\theta(x_j)}{\|E_\theta(x_i)\|\cdot\|E_\theta(x_j)\|} = \frac{r_i^\top r_j}{\|r_i\|\cdot\|r_j\|}.
\end{equation}

Due to the lack of the label guidance, this instance-level discrimination helps the model separate varying classes of samples while it does not concentrate them well~\cite{khosla2020supervised}. In this work, to congregate the representations that belong to the same class, we leverage both the label information and the strong data augmentation as the basis for constructing positive and negative pairs, as inspired by~\cite{khosla2020supervised}. The reason why we borrow the aggressive data augmentation in the self-supervised contrastive learning is two-fold: firstly, as in the traditional classification task, it increases the amount of training data and broaden the decision boundary of the classes, which helps to smooth the model and improves its generalization ability; secondly, as an additional benefit for CIL, it somewhat allows the model to see some of the past data since the high-resolution images share some similar low-resolution patches. When the model is trained to contrast these randomly cropped patches, it retains the classification ability due to those patches shared across phases. We denote the positive set $P(x_i)$ for given $x_i$ and augmentation distribution $\mathcal{A}$ by the union of the augmented and the same-class samples,
\begin{equation}
    P(x_i) = \left\{ \alpha (x_i) | \alpha \sim \mathcal{A} \right\} \cup \left\{ x_j | y_j = y_i, i \neq j  \right\}.
\end{equation}

In general, we write out the loss function of the Contrastive Class Concentration at training phase $t$, 
\begin{align}
\mathcal{L}^{(t)}_{\text{con}} &= {E}_{\substack{x_i \sim D^{(t*)}_a\\ x_p \sim P(x_i)} }\left[-\log\frac{\operatorname{exp}(s(x_i,x_p))}{{E}_{x_d\sim D^{(t*)}_a } \left[ \operatorname{exp}(s(x_i,x_d))\right]}\right],
\end{align}
where $D^{(t*)}$ is the union of the dataset of the phase $t$ and the memory bank as introduced before, and $D^{(t*)}_a$ is the union of $D^{(t*)}$ and its augmentation data.

\subsection{Representation-Level Distillation}
\label{section3.3}
Different from the classification-level distillation that restrains the distance between the probability distribution of  $E_\theta^{(t)}(X^{(t)})$ and $E_\theta^{(t-1)}(X^{(t)})$ on the previous classification heads $[W^{(1)}, \cdots, W^{(t-1)}]$, the Representation-Level Distillation~(RLD) loss we propose aims to further keep the new classes' representation distribution static during the training. 
When new-class data is input to the old model $E_\theta^{(t-1)}$, due to the congregation effect of the proposed Contrastive Class Concentration, the representation distribution of the new classes is expected to be non-overlapping over other classes. This provides a warm start for the model to allocate the new-data's representation. Therefore, explicitly constraining it to be static helps the model learn faster. 
The RLD loss $\rld ^{(t)}$ is defined as: 



\begin{figure*}[t]
	\centering
	\begin{subfigure}[t]{0.25\linewidth}
	\centering
	\includegraphics[width=1.62in]{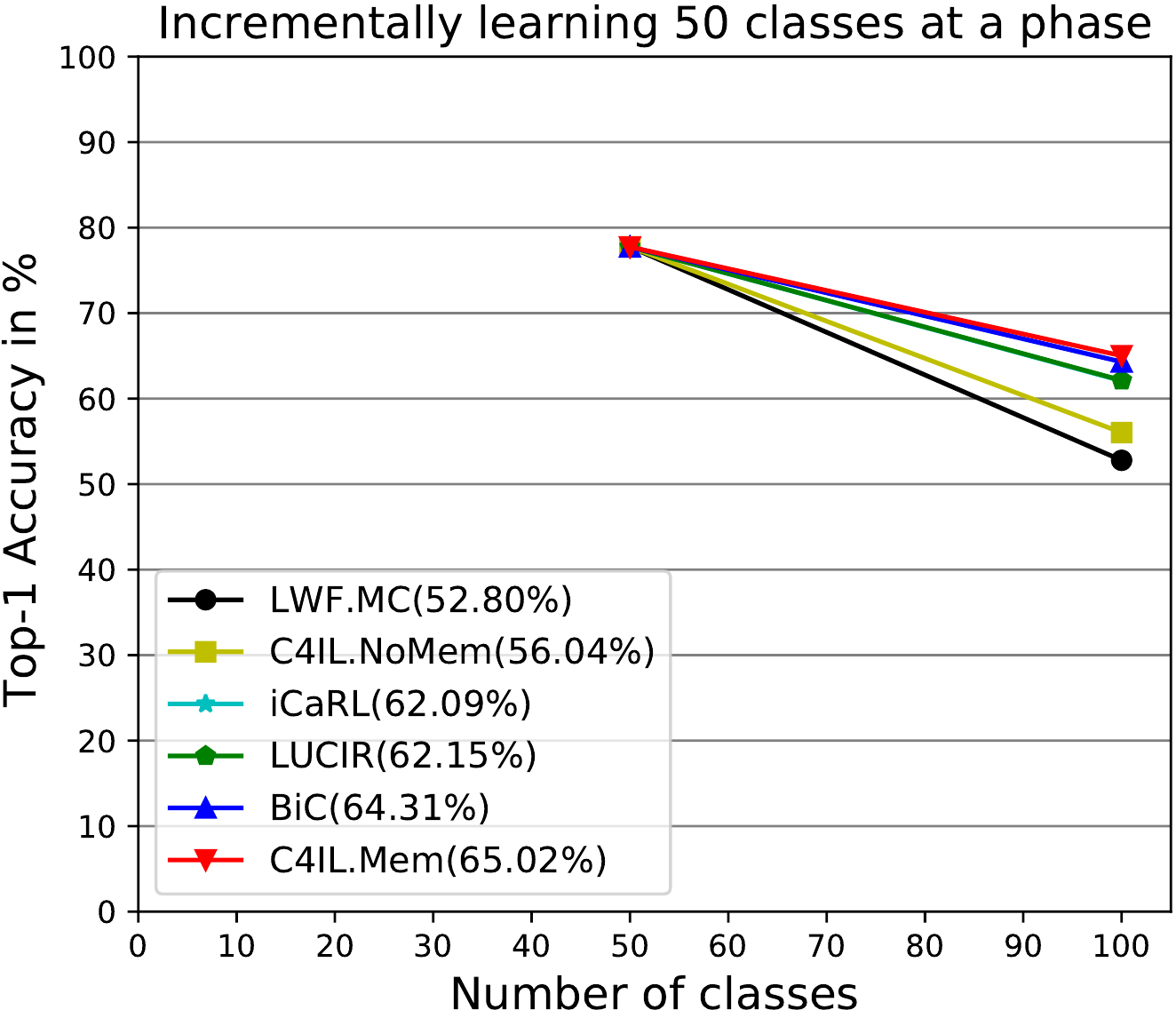}
	\end{subfigure}%
	\begin{subfigure}[t]{0.25\linewidth}
	\centering
	\includegraphics[width=1.62in]{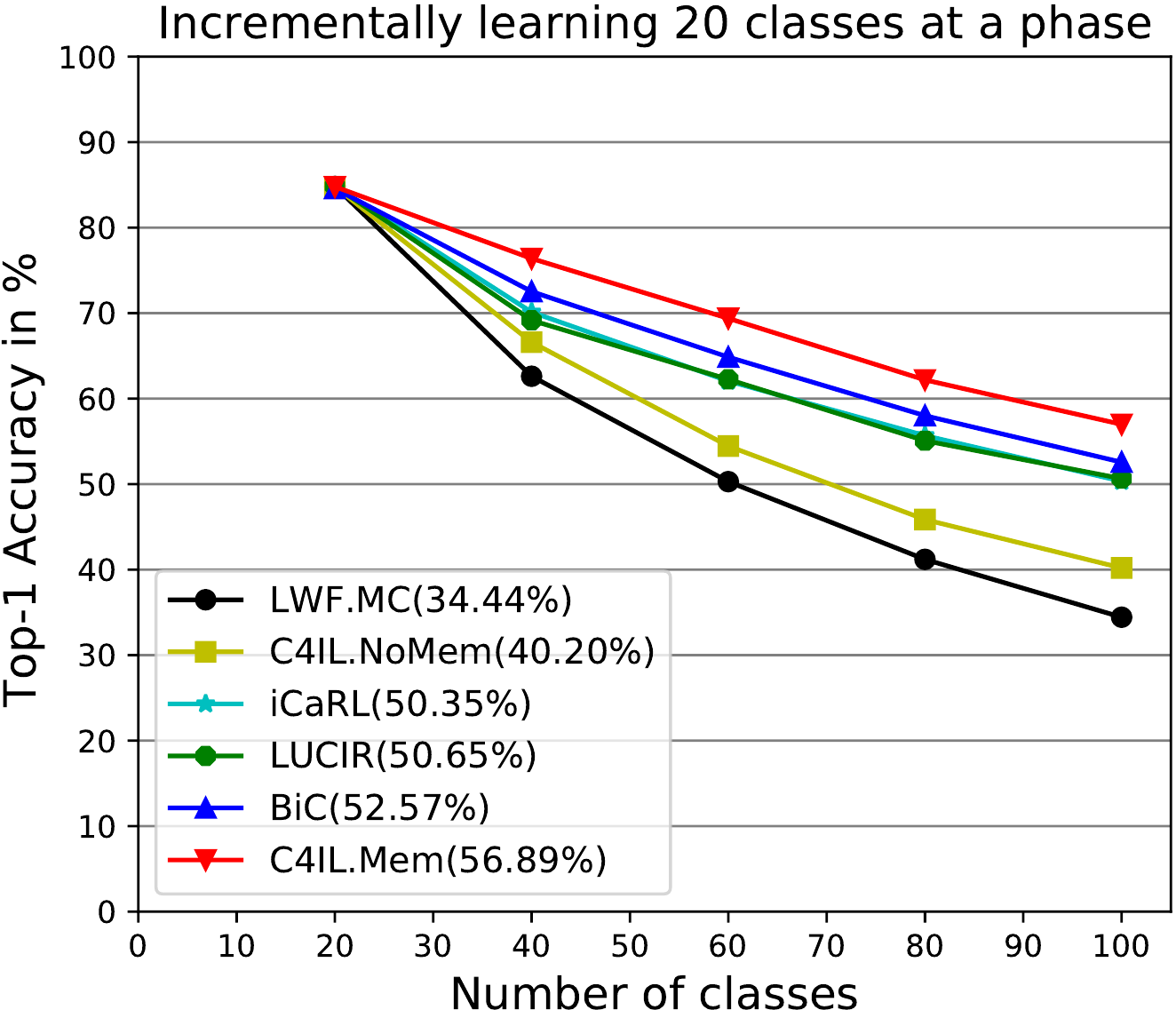}
	\end{subfigure}%
	\begin{subfigure}[t]{0.25\linewidth}
	\centering
	\includegraphics[width=1.62in]{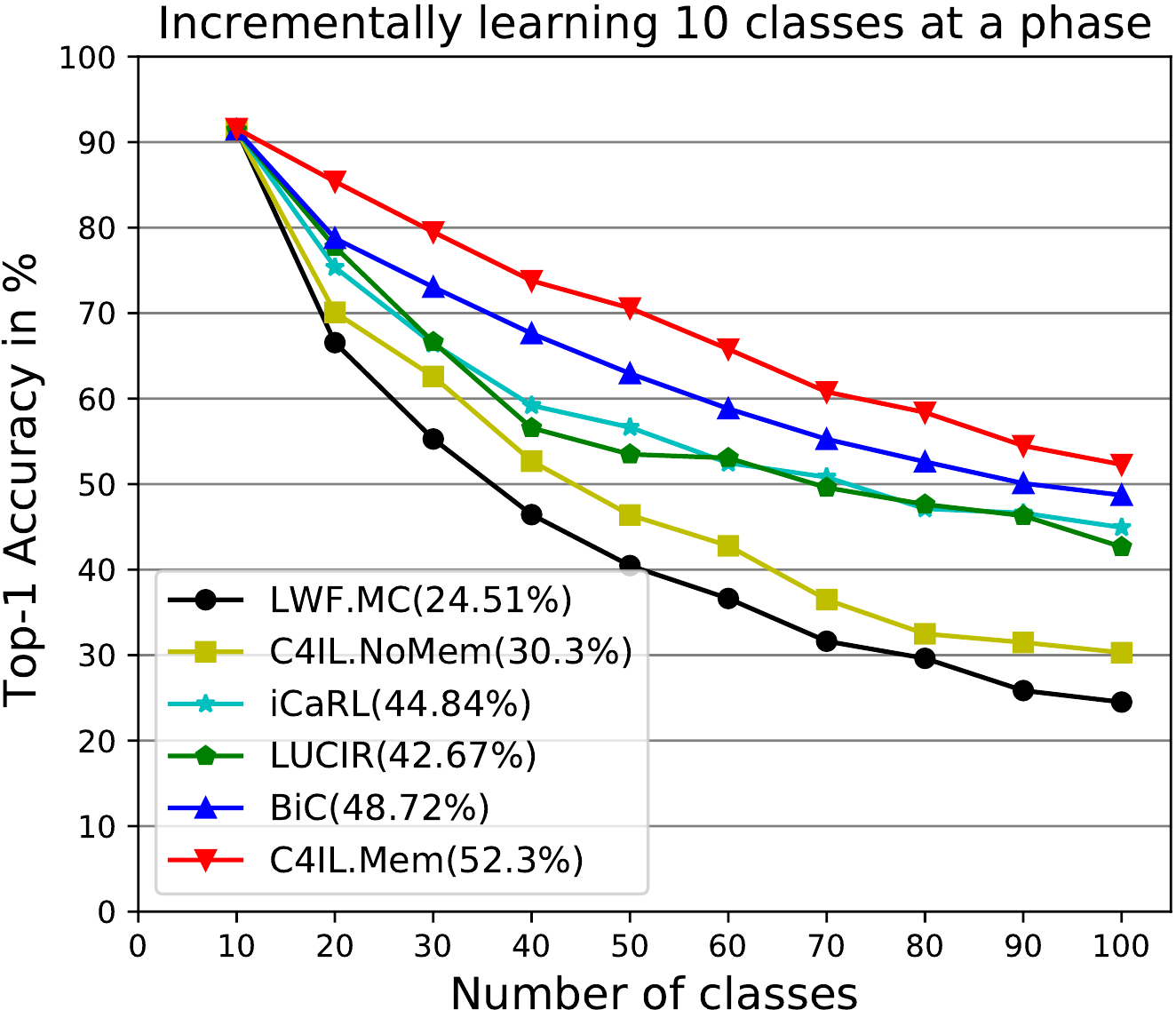}
	\end{subfigure}%
	\begin{subfigure}[t]{0.25\linewidth}
	\centering
	\includegraphics[width=1.62in]{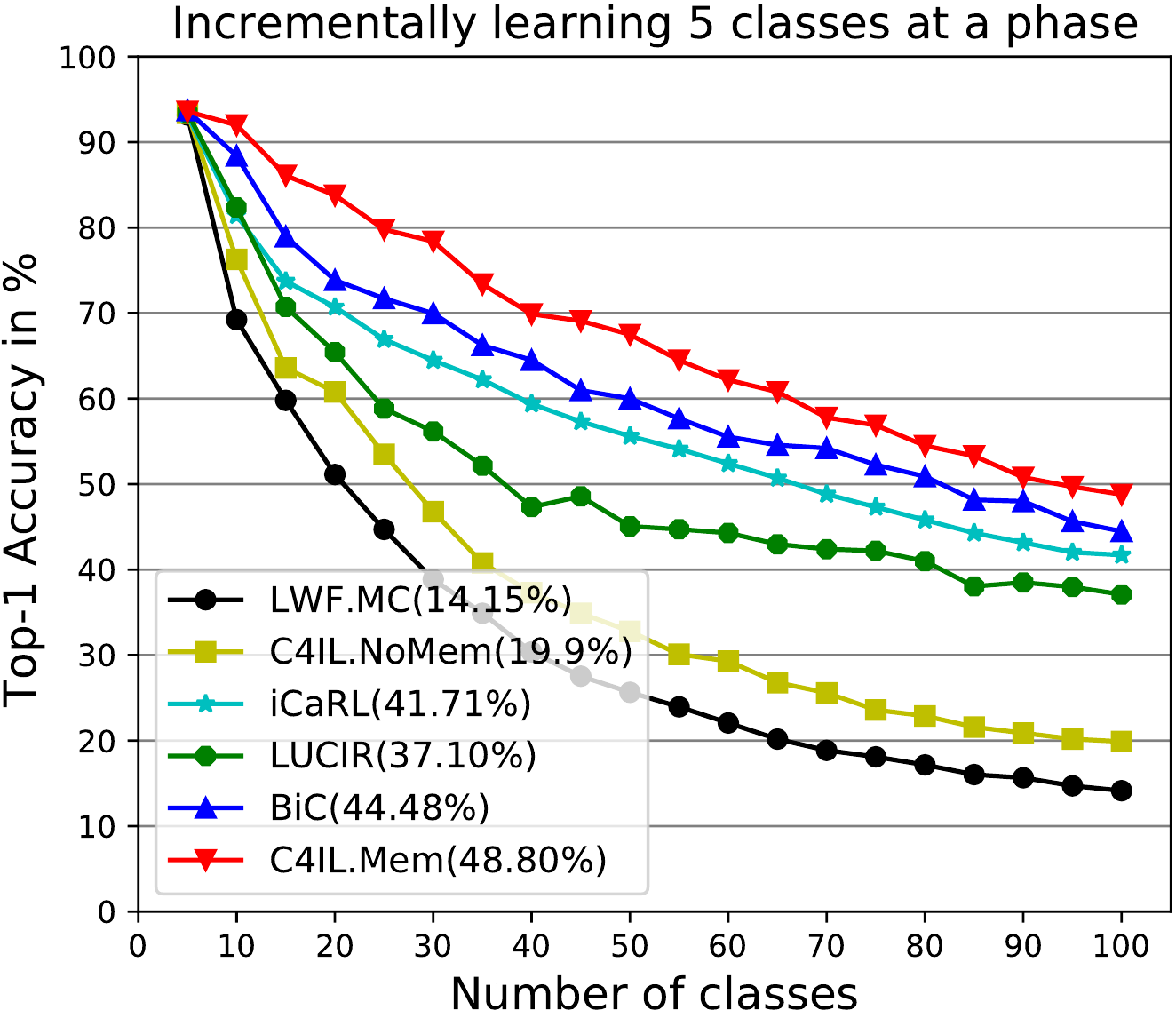}
	\end{subfigure}%
	\centering
	\caption{The top-1 accuracy of CIL compared with other methods on CIFAR-100 in batches of $50,20,10,5$ classes at a time. The final-phase top-1 accuracy is shown in the parentheses for each method, which computes over all the incremental phases except the first. Our C4IL method obtains the best results in all cases.}
	\label{compare}
	\vspace{-.3em}
\end{figure*}

\begin{equation}
\rld^{(t)} = \sum_{i \in D^{(t*)}} \left\| \gamma_i^{(t)}-\gamma_i^{(t-1)} \right\|^2,
\end{equation}
where $\gamma_i^{(t)}=E_\theta^{(t)}(x_i^{(t)})/\|E_\theta^{(t)}(x_i^{(t)})\|$ is the normalized representation sample produced by the current model $E_\theta^{(t)}$. The RLD loss normalizes the representation to a unit hypersphere, which conforms to the setting of our Contrastive Class Concentration loss in the previous sections. We argue that the RLD loss helps the model learn a consistent representation space at the current phase without forgetting the previous knowledge. Empirical findings in the following sessions also illustrate the importance of the RLD loss.

\subsection{Combining Training Objectives}
\noindent\textbf{Classificatiaon-level distillation.} Apart from concentrating representation and preserving representation distribution from the previous model, we leverage the knowledge distillation loss $\lkd^{(t)}$ at phase $t$ to retain the distribution of the soft labels, 
\begin{equation}
\lkd^{(t)} = \sum_{i=1}^{n^{(t)}}\operatorname{MSE}(\hat y_i^{(t)},\hat y_i^{(t-1)}),
\end{equation}
where $\hat y_i^{(t)}$ and $\hat y_i^{(t-1)}$ is the current and previous model's probability distribution of sample $x_i^{(t)}$ at training phase $t$.\\

\noindent\textbf{Classification loss.} In order to classify the current-phase data, we adopt the classification loss $\lce^{(t)}$:
\begin{equation}
\lce^{(t)} = \sum_{i=1}^{n^{(t)}}\operatorname{CE}(\hat y_i^{(t)},y^{(t)}_i),
\end{equation}
where the $y_i^{(t)}$ is the one-hot ground-truth label of sample $x_i^{(t)}$ and $\operatorname{CE(\cdot,\cdot)}$ stands for the cross-entropy loss function. For the memory-based method, we simply concatenate the memory bank and the current data for cross-entropy loss. \\

\noindent\textbf{Combination of the objectives.} 
Following the previous work~\cite{li2017learning,hou2019learning}, we combine all of this loss function to construct the total training objective of our model: 
\begin{equation}
\mathcal{L}^{(t)} = \lce^{(t)}+ \beta_t \mathcal{L}_{\text{con}}^{(t)}+ \kappa_t \lkd^{(t)} + \eta_t \rld^{(t)},
\end{equation}
where the $\beta_t$, $\kappa_t$ and $\eta_t$ are all functions of the phase number $t$ and update themselves as $t$ increases. For example, $\beta_t = \beta_{t-1} + \lambda (t-1)$. The details are listed in the supplementary materials. In summary, Fig.\ref{training_process} illustrates the process of C4IL at phase $t$.

\begin{figure}[!t]
\vspace{-1.5em}
	\centering
	\includegraphics[width=.49\linewidth]{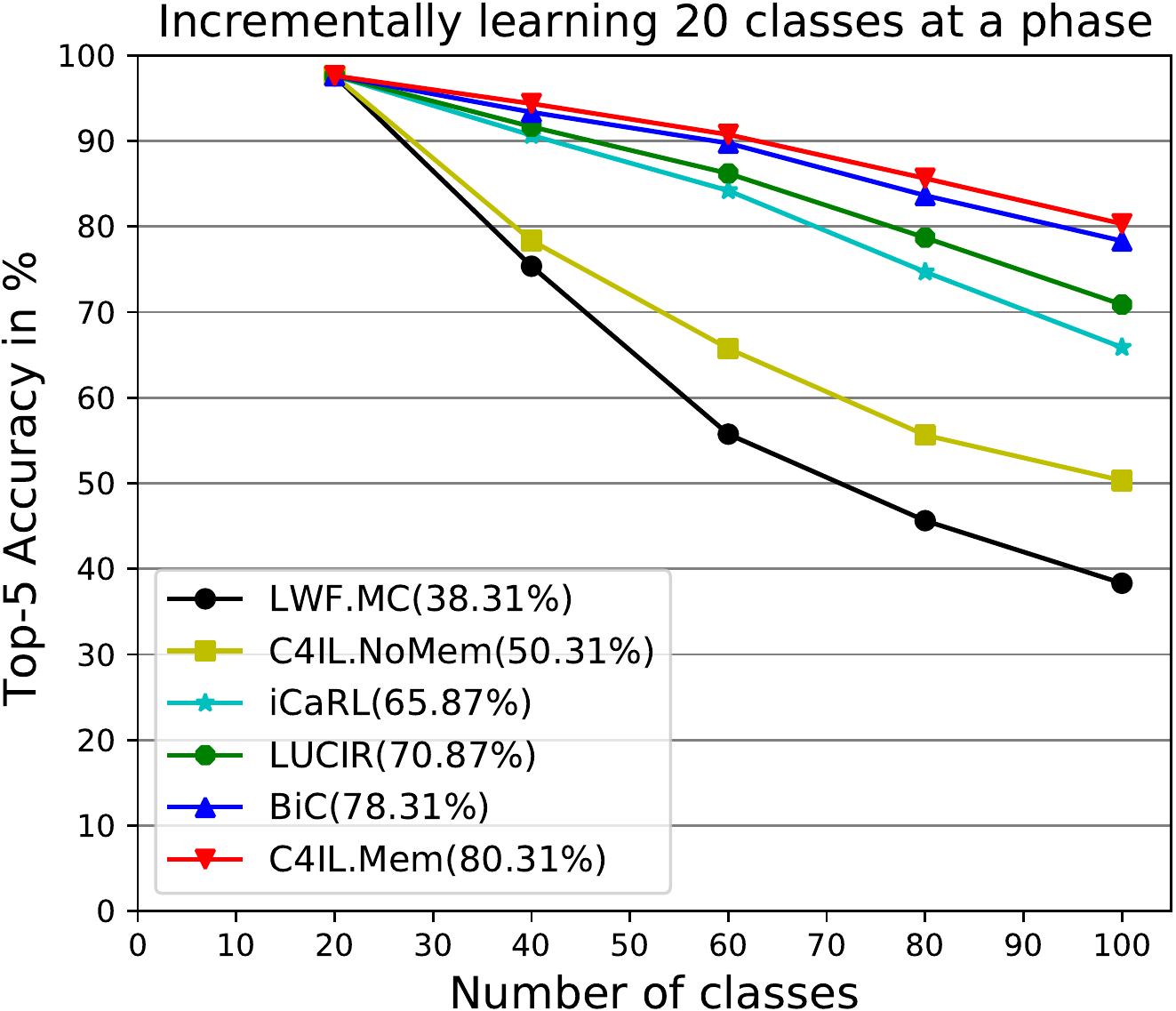}
	\includegraphics[width=.49\linewidth]{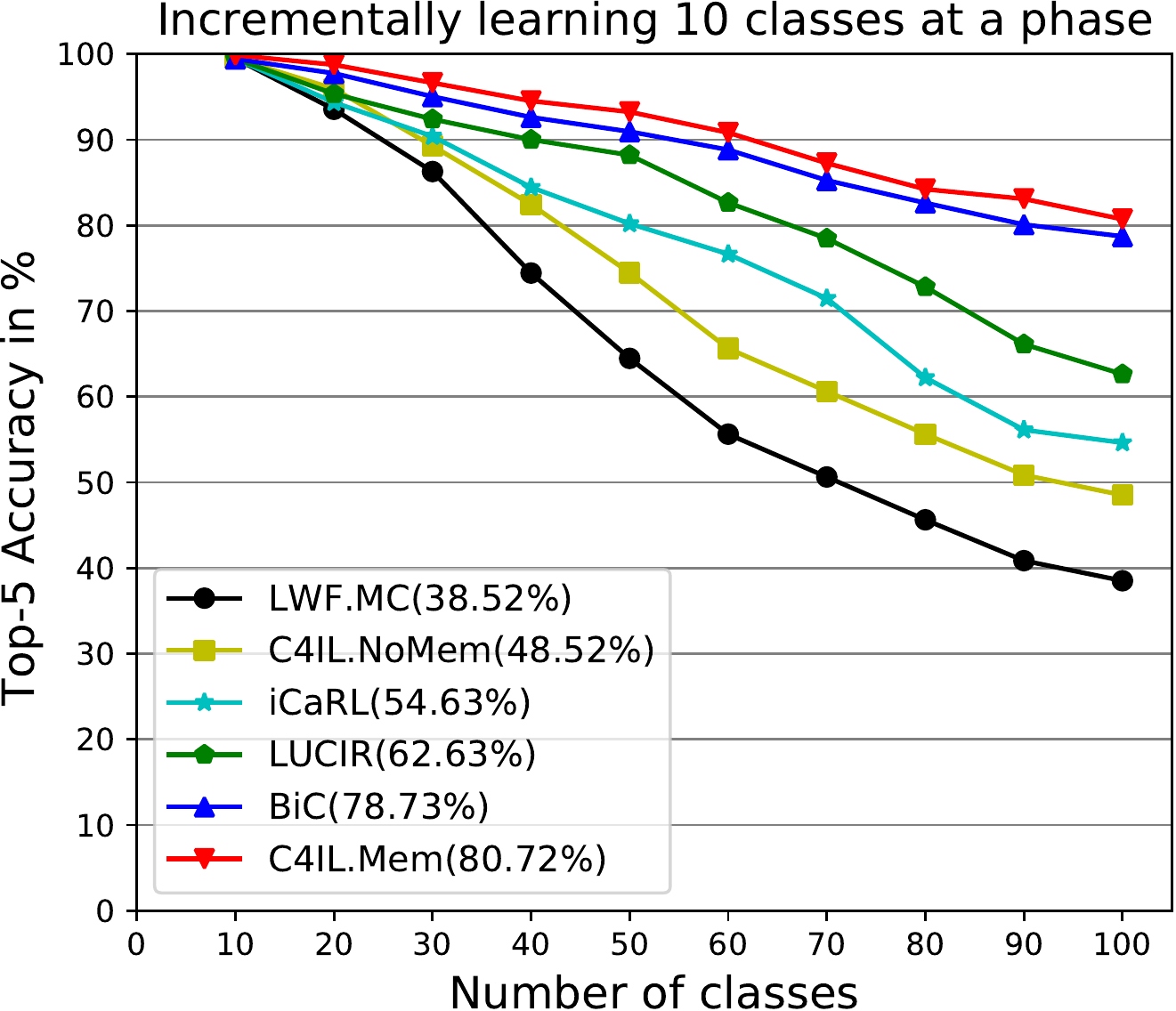}
\vspace{-.5em}
\caption{The top-5 accuracy of methods evaluated on ImageNet-100 in batches of 20 and 10 classes at a time.}
\label{fig:average_imagenet}
\vspace{-1.2em}
\end{figure}

\section{Experiments}
\subsection{Setup}
\noindent{\bf Datasets and metrics.}\quad We adopt CIFAR-100 \cite{krizhevsky2009learning} and ImageNet-100~\cite{tian2019contrastive} to benchmark the main CIL results for all methods. 
We follow the two widely adopted metrics in CIL: final accuracy~\cite{rebuffi2017icarl} and average accuracy except the first phase~\cite{hou2019learning}. The first measures the general performance of a model after CIL, and the second eliminates the influence of the first-phase performance by averaging the model's accuracies except the first phase. In addition, we introduce the linear evaluation accuracy~\cite{zhang2016colorful,chen2020simple,he2020momentum} to explicitly measure the inter-phase confusion.\\

\noindent\textbf{Data augmentation.}\quad We use a subset of image augmentations as proposed in SimCLR \cite{chen2020simple}. First, a series of spatial transformations are adopted: a random crop of the image with $\operatorname{scale}=[0.2,1]$ is selected and resized to 32$\times $32 for CIFAR-100 and 224$\times$224 for Imagenet-100 with a random horizontal flip. Secondly, a sequence of the color transformations are sampled: brightness$(0.4)$, contrast$(0.4)$, saturation$(0.4)$, hue adjustments$(0.1)$, and an optional grayscale$(p=0.2)$.\\ 

\noindent\textbf{Class incremental setting.}\quad Two popular strategies of simulating class incremental learning are often adopted. The first~\cite{rebuffi2017icarl,wu2019large,kirkpatrick2017overcoming,castro2018end} evenly splits the dataset into $T$ phases, and the model is trained sequentially. The second~\cite{hou2019learning,hu2021distilling} uses half of the data to pre-train a model and splits the remaining data into $T$ phases. Due to the large amount of data for pre-training, the second strategy often achieves a high baseline for subsequent CIL and weakens the impact of catastrophic forgetting. Therefore in this paper, we adopt the first strategy since it is more challenging and helps to highlight the model's efficacy of addressing catastrophic forgetting.

As the previous CIL work \cite{rebuffi2017icarl,wu2019large,kirkpatrick2017overcoming,castro2018end}, we split CIFAR-100 dataset into $T=2,5,10,20$ incremental phases, and each sub-dataset has $50000/T$ training data and $100/T$ class labels. Using a 32-layers ResNet as encoder and evaluate its Top-1 accuracy in each phase. For ImageNet-100, we use an 18-layers ResNet as backbone and evaluate its Top-5 accuracy in $T=5,10$ incremental phases which is consistent with previous work~\cite{rebuffi2017icarl,wu2019large,kirkpatrick2017overcoming,castro2018end}.
As for the combination of the different losses, we use the grid search to find the optimal weights for different losses. Please refer to the supplementary materials for the detail.
To better illustrate the effect of our method, we respectively compare the results with and without a memory bank. In memory bank case, we consider a memory with fixed capacity $|\dmem| = 2000$ which is consistent with the previous works \cite{rebuffi2017icarl,wu2019large,hou2019learning}. Since the capacity is independent of the number of classes, when the number of stored classes increases, we randomly discard a fixed number of samples in each class to keep on the capacity $|\dmem|$ is always 2000. So that the more classes stored, the fewer samples are reserved for each old class. 

\begin{table}[t]
  \small
  \tablestyle{2.5pt}{1.1}
  \begin{tabular}{cccccccc}
  \toprule
  & \multicolumn{4}{c}{\textbf{CIFAR-100}} &  & \multicolumn{2}{c}{\textbf{ImageNet-100}}  
  		\\ \cline{2-5} \cline{7-8} 
   
  \multirow{-2}{*}{\quad \textbf{Methods} \quad} & {\textit{P=2}} & {\textit{5}} & {\textit{10}} & {\textit{20}} & \textbf{}&{\textit{5}} & {\textit{10}} \\ \hline\hline
   
  \smaller{LwF\cite{li2017learning}} & 52.76 & 47.15 & 39.89 & 29.64 &  & 53.75 & 61.12  \\
  \smaller{\textbf{C4IL.NoMem}} & \textbf{56.04} & \textbf{51.79} & \textbf{44.04} & \textbf{34.14} & & \textbf{62.5} & \textbf{69.27} \\
  \hline
  \smaller{iCaRL\cite{rebuffi2017icarl}} & 62.09 & 59.55 & 55.51 & 55. 91 & & 78.85 & 74.50 \\
  \smaller{LUCIR\cite{hou2019learning}} & 62.15 & 59.29 & 54.86 & 49.26 & & 81.85 & 80.96 \\
  \smaller{BiC\cite{wu2019large}} & 64.31 & 62 & 60.88 & 60.32 & & 86.25 & 87.93 \\
  \smaller{\textbf{C4IL.Mem}} & \textbf{65.02} & \textbf{66.25} & \textbf{66.79} & \textbf{66.28} & & \textbf{87.75} & \textbf{89.92}\\
  \bottomrule
  \end{tabular}
    \caption{Average accuracy of CIL on CIFAR-100 and ImageNet-100. Our C4IL obtains the best results in all cases.}
\vspace{-0.3em}
\label{average_result}
\end{table}

\subsection{Comparison to the SOTA Methods}
In this section, we evaluate our method's performance by comparing it with other state-of-the-art models. For the memory-free baseline, we use the classic LwF.MC~\cite{li2017learning} as its distillation technique is widely adopted by existing methods. For the memory-based baseline, we first include iCaRL~\cite{rebuffi2017icarl} since it's the first well recognized method that has thoroughly discussed multiple aspects of the memory in CIL. We also include BiC~\cite{wu2019large}, a reproducible framework with the best performance among the baselines. We further include LUCIR~\cite{hou2019learning} since it is another framework that takes explicit measures that have effect on alleviating inter-phase confusion. 

In Fig.\ref{compare} and Fig.\ref{fig:average_imagenet}, each curve illustrates the method's top-1/5 accuracy for all the past classes at the current training phase, and the accuracies after the whole CIL training are shown in the parentheses. 
The proposed method C4IL with the memory bank (C4IL.Mem) outperforms all the baselines by a significant margin across all the settings consistently. It achieve higher accuracy at each learning phase and when the new-class data are added, its accuracy decreases much more slowly compared to the baselines. 
In the cases where no memory bank is adopted, our method C4IL without memory (C4IL.NoMem) is superior to LwF.MC in every CIL setting. 
As summarized in Tab.\ref{average_result}, the average accuracy of C4IL outperforms the baselines by a large margin (4\% and 6\% ) with and without memory, especially when the number of phases is large. Our method C4IL is significantly higher than other methods in CIFAR-100 (6\%) and ImageNet-100 (8\%) regardless of the existence of the memory bank.

\begin{figure}[t]
	\begin{subfigure}[t]{1\linewidth}
	\centering
	\includegraphics[width=3.2in]{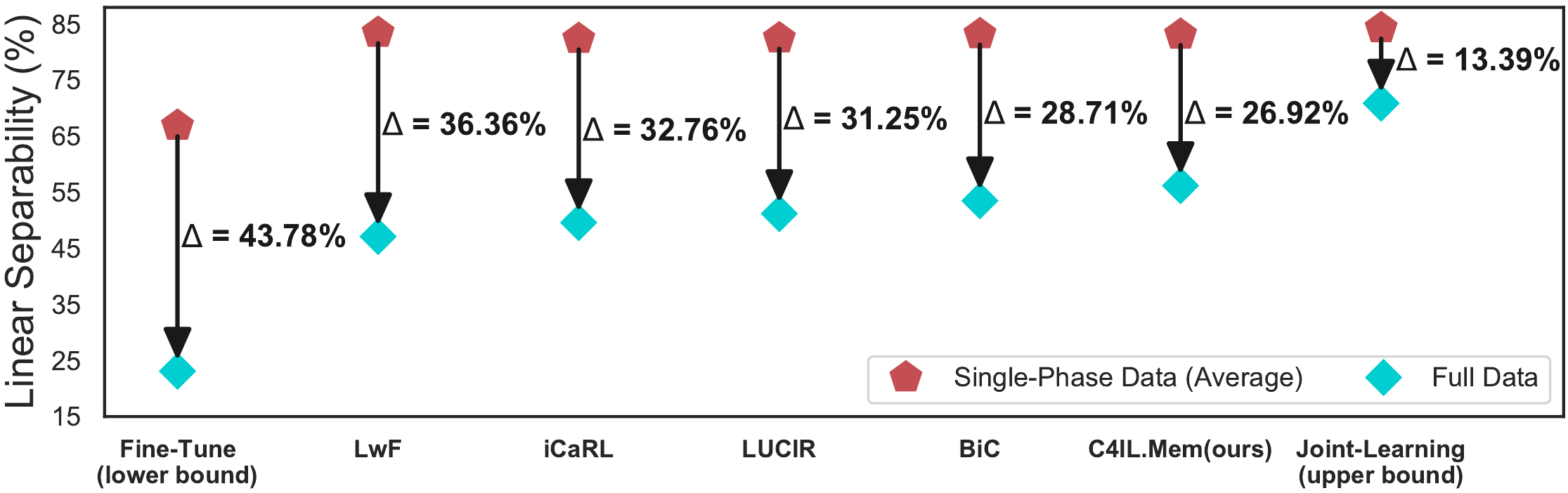}
	\vspace{0.1em}
	\caption{Study of Inter-phase Confusion of Existing Methods.}
	\label{fig:inter-phase-ours}
	\end{subfigure}%
	\vspace{0.8em}
	\begin{subfigure}[t]{0.5\linewidth}
	\centering
	\includegraphics[width=1.6in]{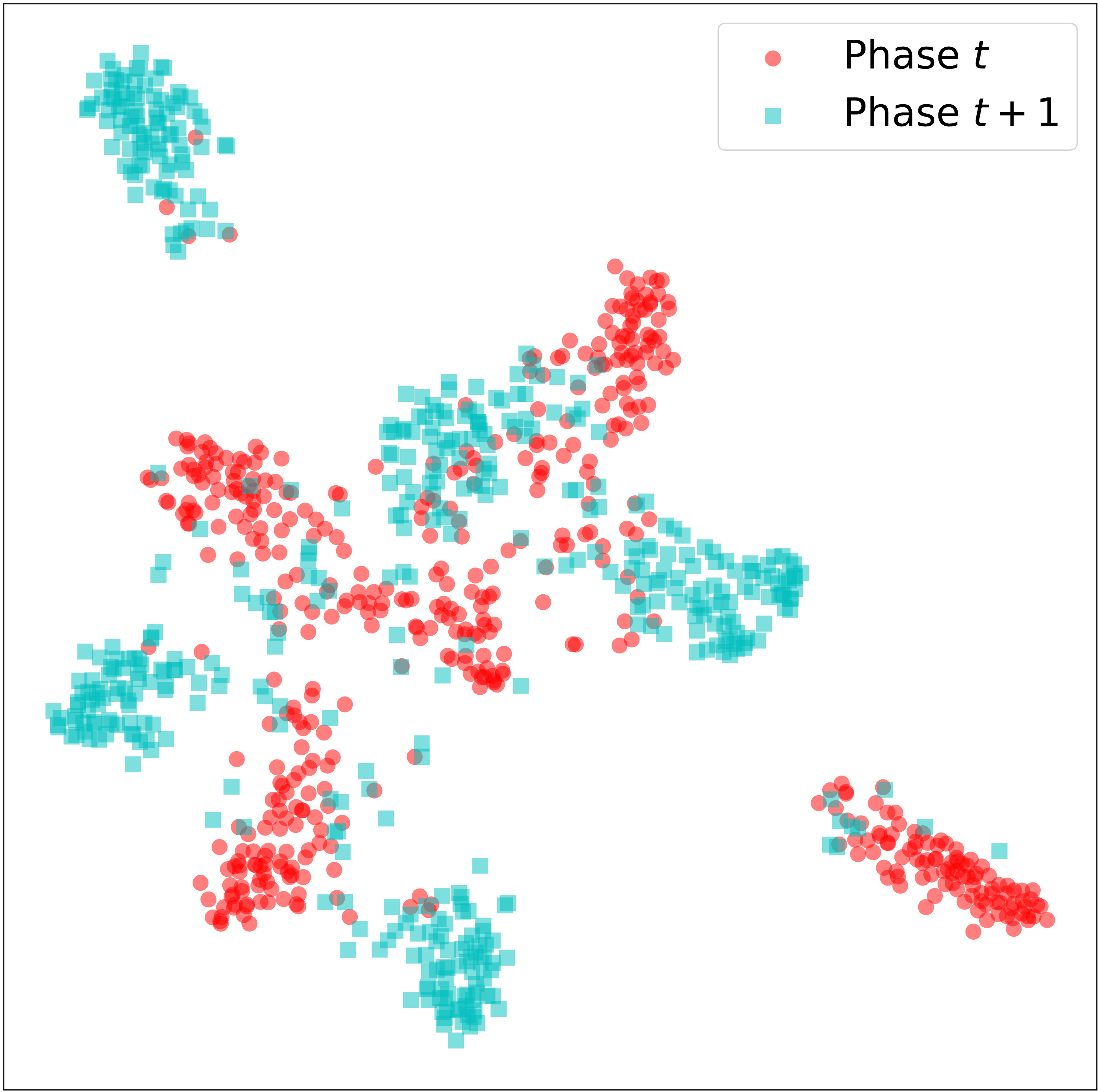}
	\caption{t-SNE of iCaRL}
	\label{fig:tsne-icarl}
	\end{subfigure}%
	\begin{subfigure}[t]{0.5\linewidth}
	\centering
	\includegraphics[width=1.6in]{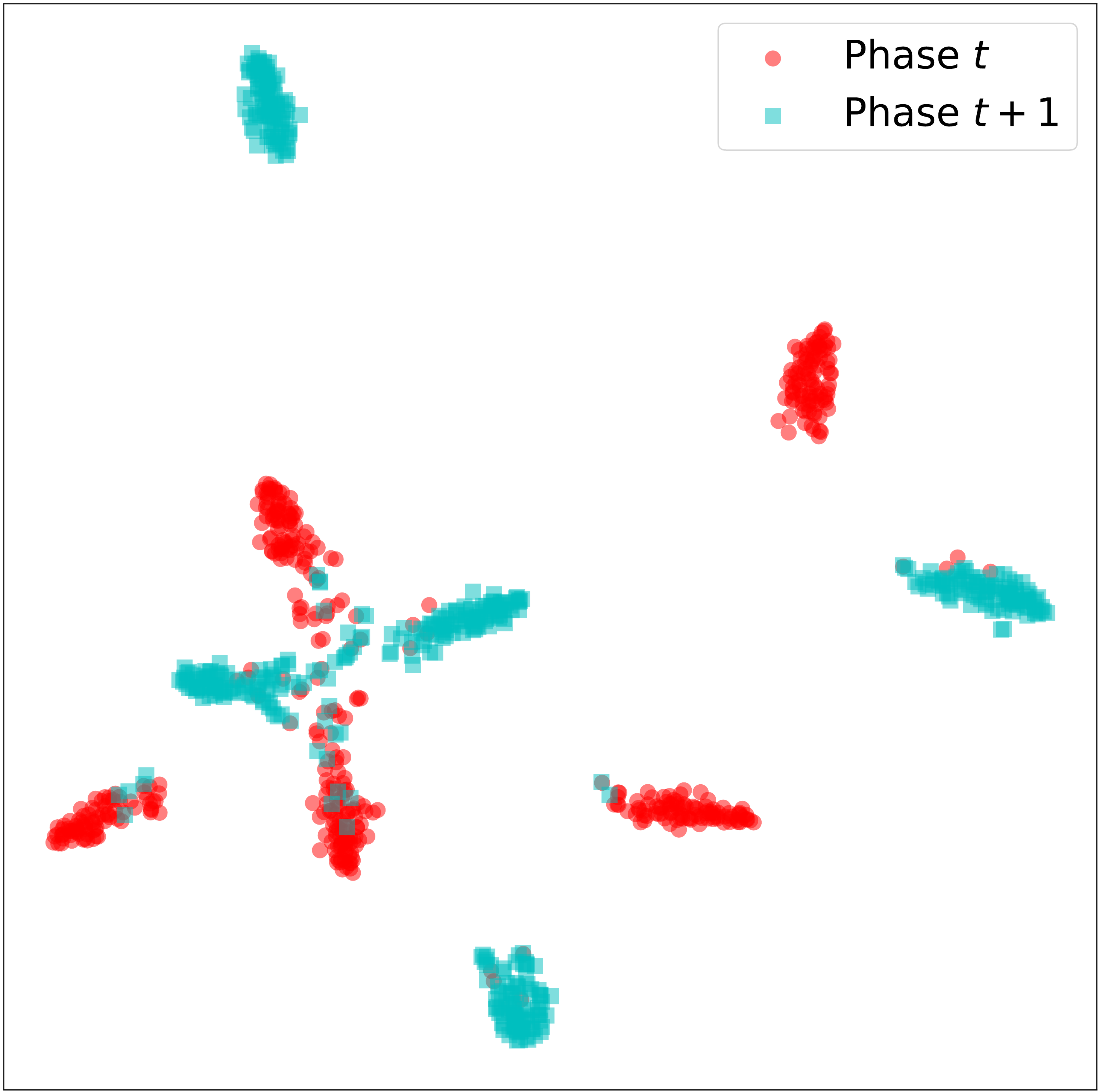}
	\caption{t-SNE of C4IL.Mem}
	\label{fig:tsne-c4il}
	\end{subfigure}%
	\centering
	\caption{Verification of C4IL's ability of alleviating inter-phase confusion, evaluated on CIFAR-100~\cite{krizhevsky2009learning}.
	\textbf{(a)}: Differences between the average linear separability of single-phase data and the linear separability of full data. C4IL successfully alleviates inter-phase confusion with the best performance on full data and the smallest $\Delta$.
	\textbf{(b)}\&\textbf{(c)}: The t-SNE visualization of representation distribution at two successive phases (5 classes per phase, 10 classes in total).
	It shows that compared to iCaRL, our method C4IL.Mem suffers less from inter-phase confusion.
	}
	\vspace{-1.3em}
	\label{ROshow}
\end{figure}

\subsection{C4IL Alleviates Inter-phase Confusion}
To verify that the performance gain of C4IL is yielded by the effort of alleviating inter-phase confusion, we adopt both quantitative and qualitative demonstrations.

Quantitatively, inn Fig.\ref{fig:inter-phase-ours}, we first evaluate the linear separability of the baselines and our method on the data of each individual phase. Then we compare it with the linear separability of the full data. The difference between the two accuracies are composed of two factors: the inevitable intrinsic data overlapping~(as in joint learning upper bound) and inter-phase confusion. C4IL achieves the best linear separability on the full data~(56.20\%) and the smallest performance difference~(26.92\%), which confirms that C4IL indeed addresses the problem of inter-phase confusion.

Qualitatively, in Fig.\ref{fig:tsne-icarl} and Fig.\ref{fig:tsne-c4il}, we use t-SNE~\cite{maaten2008visualizing} to visualize the representation distribution of iCaRL and C4IL.Mem in two successive phases. The red circles denote the phase-$t$ data representations generated by the current phase-$t+1$ model. The blue squares are the phase-$t+1$ representations. They demonstrate that C4IL does concentrate the representations and therefore alleviate inter-phase confusion.

\begin{table}[t]
\footnotesize
\centering
\begin{tabular}{lccc}
\shline
Methods	& Linear Acc (\%)& Final Acc (\%)  & Avg Acc (\%) \\
\hline
\hline
C4IL.Mem	& \textbf{56.24} &	\textbf{52.30}		&	\textbf{66.79} \\ 
\quad $-$ DA & 52.71(-3.53)		&	49.40~(-2.90)	&	62.01~(-4.78) \\
\quad $-$ RLD   & 53.42(-2.82)   &   50.60~(-1.70)      &   64.33~(-2.46)\\
\quad $-$ LG  & 51.48(-4.76)  &   46.37~(-5.93)     &   61.64~(-5.15) \\ 
\hline
iCaRL	& \textbf{49.62} & \textbf{44.84}	& \textbf{55.51} \\
\quad $+$ DA  & 45.23(-4.39) &   41.21~(-3.63)  &   50.24~(-5.27)\\
\shline
\end{tabular}
\vspace{-.3em}
\caption{C4IL ablative study, evaluated on CIFAR-100~\cite{krizhevsky2009learning}. 
\textbf{DA}: aggressive data augmentation;
\textbf{RLD}: representation-level distillation;
\textbf{LG}: label guidance.
}
\vspace{-1.2em}
\label{tab:ablation}
\end{table}

\subsection{Ablative Studies}
To provide more in-depth insights into the working mechanism of C4IL, we conduct ablative studies on three important ingredients: Data Augmentation~(DA), Representation-Level Distillation~(RLD), and Label Guidance~(LG). As shown in Tab.\ref{tab:ablation}, we base our study on phase-10 CIL training on CIFAR-100, and evaluate the variants on three metrics: linear evaluation accuracy, final accuracy, and average accuracy except the first phase. The first metric helps to understand the method's effectiveness of alleviating inter-phase confusion, and the rest two measure the general capability against catastrophic forgetting.\\

\noindent\textbf{Data Augmentation~(DA).}\quad 
The aggressive data augmentation is beneficial to C4IL for alleviating inter-phase confusion. Without it, C4IL still gains a small amount of improvement over iCaRL, in terms of both linear separability~(56.24 v.s 49.62) and final accuracy~(52.30 v.s 44.84). However, the benefit of DA is not universally applicable and its usage should be accompanied by the contrastive learning paradigm: iCaRL has a severe performance decrease when DA is applied~(49.62\ding{213}45.23). This is because the strong distortion impairs the connection between the labels and the original-domain images, weakening the discriminative power of simple image classification frameworks~\cite{goodfellow2014explaining,chen2020simple}. Therefore, we conclude that the data augmentation of random cropping and color distortion is significantly beneficial to alleviating inter-phase confusion but should be used with extra care.\\


\noindent\textbf{Representation-Level Distillation~(RLD).}\quad 
RLD is designed to maintain the shape of the representation distribution of the previous models. As shown in Tab.\ref{tab:ablation}, RLD loss is instrumental to the performance of C4IL, but it does not play the most critical role in C4IL: the variant without RLD has the smallest performance drop evaluated on all three metrics. We conjecture that this is because the old class data in memory bank also provides a certain constraint on the representation distribution but indirectly, which is somewhat similar to the effect of the RLD loss.\\


\noindent\textbf{Label Guidance~(LG).}\quad
It has been validated that when the contrastive learning objective is guided by the label, the generated representation space is more compact~\cite{khosla2020supervised}. As shown in Tab.\ref{tab:ablation}, we also observe that the label guidance is the most important component in C4IL, playing a crucial role to concentrate the same-class data and lower the probability of inter-phase confusion. Ablating it will cause the most severe performance drops on all three metrics.
\section{Conclusion}
By introducing the measure of representation quality into Class Incremental Learning~(CIL) for the first time, three causes for catastrophic forgetting are decoupled and systematically studied in this paper. Among them, the cause of ``inter-phase forgetting'' is a crucial performance bottleneck and needs to be further explored. Then, a straightforward measure C4IL is proposed towards this specific issue, the empirical effectiveness of which highlights the importance of the problem of inter-phase confusion. We hope the representation learning viewpoint we take in this paper can provide some insights for the future research. \\


\noindent\textbf{Societal Impact}: this work contributes to fundamental research without any societal impact.

\noindent\textbf{Limitations}: our exploration is limited by computational resources and cannot carry out verification on other larger-scale datasets such as ImageNet1k~\cite{russakovsky2015imagenet}. 
We didn't discuss memory bank in terms of its efficacy of addressing the three causes for forgetting. This is mainly because the memory bank itself is a relaxation~(or violation?) of CIL setting and apparently approaches all three causes simultaneously. 

{\small
\clearpage
\newpage
\newpage
\bibliographystyle{ieee_fullname}
\bibliography{egbib}

\begin{thebibliography}{10}\itemsep=-1pt

\bibitem{benzing2020understanding}
Frederik Benzing.
\newblock Understanding regularisation methods for continual learning.
\newblock {\em arXiv e-prints}, pages arXiv--2006, 2020.

\bibitem{castro2018end}
Francisco~M Castro, Manuel~J Mar{\'\i}n-Jim{\'e}nez, Nicol{\'a}s Guil, Cordelia
  Schmid, and Karteek Alahari.
\newblock End-to-end incremental learning.
\newblock In {\em Proceedings of the European conference on computer vision
  (ECCV)}, pages 233--248, 2018.

\bibitem{chen2020simple}
Ting Chen, Simon Kornblith, Mohammad Norouzi, and Geoffrey Hinton.
\newblock A simple framework for contrastive learning of visual
  representations.
\newblock {\em arXiv preprint arXiv:2002.05709}, 2020.

\bibitem{douillard2020podnet}
Arthur Douillard, Matthieu Cord, Charles Ollion, Thomas Robert, and Eduardo
  Valle.
\newblock Podnet: Pooled outputs distillation for small-tasks incremental
  learning.
\newblock In {\em Computer Vision--ECCV 2020: 16th European Conference,
  Glasgow, UK, August 23--28, 2020, Proceedings, Part XX 16}, pages 86--102.
  Springer, 2020.

\bibitem{goodfellow2013empirical}
Ian~J Goodfellow, Mehdi Mirza, Da Xiao, Aaron Courville, and Yoshua Bengio.
\newblock An empirical investigation of catastrophic forgetting in
  gradient-based neural networks.
\newblock {\em arXiv preprint arXiv:1312.6211}, 2013.

\bibitem{goodfellow2014explaining}
Ian~J Goodfellow, Jonathon Shlens, and Christian Szegedy.
\newblock Explaining and harnessing adversarial examples.
\newblock {\em arXiv preprint arXiv:1412.6572}, 2014.

\bibitem{he2018exemplar}
Chen He, Ruiping Wang, Shiguang Shan, and Xilin Chen.
\newblock Exemplar-supported generative reproduction for class incremental
  learning.
\newblock In {\em BMVC}, page~98, 2018.

\bibitem{he2020incremental}
Jiangpeng He, Runyu Mao, Zeman Shao, and Fengqing Zhu.
\newblock Incremental learning in online scenario.
\newblock In {\em Proceedings of the IEEE/CVF Conference on Computer Vision and
  Pattern Recognition}, pages 13926--13935, 2020.

\bibitem{he2020momentum}
Kaiming He, Haoqi Fan, Yuxin Wu, Saining Xie, and Ross Girshick.
\newblock Momentum contrast for unsupervised visual representation learning.
\newblock In {\em Proceedings of the IEEE/CVF Conference on Computer Vision and
  Pattern Recognition}, pages 9729--9738, 2020.

\bibitem{hinton2015distilling}
Geoffrey Hinton, Oriol Vinyals, and Jeff Dean.
\newblock Distilling the knowledge in a neural network.
\newblock {\em arXiv preprint arXiv:1503.02531}, 2015.

\bibitem{hou2018lifelong}
Saihui Hou, Xinyu Pan, Chen Change~Loy, Zilei Wang, and Dahua Lin.
\newblock Lifelong learning via progressive distillation and retrospection.
\newblock In {\em Proceedings of the European Conference on Computer Vision
  (ECCV)}, pages 437--452, 2018.

\bibitem{hou2019learning}
Saihui Hou, Xinyu Pan, Chen~Change Loy, Zilei Wang, and Dahua Lin.
\newblock Learning a unified classifier incrementally via rebalancing.
\newblock In {\em Proceedings of the IEEE Conference on Computer Vision and
  Pattern Recognition}, pages 831--839, 2019.

\bibitem{hu2021distilling}
Xinting Hu, Kaihua Tang, Chunyan Miao, Xian-Sheng Hua, and Hanwang Zhang.
\newblock Distilling causal effect of data in class-incremental learning.
\newblock In {\em Proceedings of the IEEE/CVF Conference on Computer Vision and
  Pattern Recognition}, pages 3957--3966, 2021.

\bibitem{khosla2020supervised}
Prannay Khosla, Piotr Teterwak, Chen Wang, Aaron Sarna, Yonglong Tian, Phillip
  Isola, Aaron Maschinot, Ce Liu, and Dilip Krishnan.
\newblock Supervised contrastive learning.
\newblock {\em arXiv preprint arXiv:2004.11362}, 2020.

\bibitem{kirkpatrick2017overcoming}
James Kirkpatrick, Razvan Pascanu, Neil Rabinowitz, Joel Veness, Guillaume
  Desjardins, Andrei~A Rusu, Kieran Milan, John Quan, Tiago Ramalho, Agnieszka
  Grabska-Barwinska, et~al.
\newblock Overcoming catastrophic forgetting in neural networks.
\newblock {\em Proceedings of the national academy of sciences},
  114(13):3521--3526, 2017.

\bibitem{krizhevsky2009learning}
Alex Krizhevsky, Geoffrey Hinton, et~al.
\newblock Learning multiple layers of features from tiny images.
\newblock 2009.

\bibitem{lesort2019regularization}
Timoth{\'e}e Lesort, Andrei Stoian, and David Filliat.
\newblock Regularization shortcomings for continual learning.
\newblock {\em arXiv preprint arXiv:1912.03049}, 2019.

\bibitem{li2017learning}
Zhizhong Li and Derek Hoiem.
\newblock Learning without forgetting.
\newblock {\em IEEE transactions on pattern analysis and machine intelligence},
  40(12):2935--2947, 2017.

\bibitem{liu2020more}
Yu Liu, Sarah Parisot, Gregory Slabaugh, Xu Jia, Ales Leonardis, and Tinne
  Tuytelaars.
\newblock More classifiers, less forgetting: A generic multi-classifier
  paradigm for incremental learning.
\newblock In {\em Computer Vision--ECCV 2020: 16th European Conference,
  Glasgow, UK, August 23--28, 2020, Proceedings, Part XXVI 16}, pages 699--716.
  Springer, 2020.

\bibitem{liu2020mnemonics}
Yaoyao Liu, Yuting Su, An-An Liu, Bernt Schiele, and Qianru Sun.
\newblock Mnemonics training: Multi-class incremental learning without
  forgetting.
\newblock In {\em Proceedings of the IEEE/CVF Conference on Computer Vision and
  Pattern Recognition}, pages 12245--12254, 2020.

\bibitem{maaten2008visualizing}
Laurens van~der Maaten and Geoffrey Hinton.
\newblock Visualizing data using t-sne.
\newblock {\em Journal of machine learning research}, 9(Nov):2579--2605, 2008.

\bibitem{mittal2021essentials}
Sudhanshu Mittal, Silvio Galesso, and Thomas Brox.
\newblock Essentials for class incremental learning.
\newblock In {\em Proceedings of the IEEE/CVF Conference on Computer Vision and
  Pattern Recognition}, pages 3513--3522, 2021.

\bibitem{oord2018representation}
Aaron van~den Oord, Yazhe Li, and Oriol Vinyals.
\newblock Representation learning with contrastive predictive coding.
\newblock {\em arXiv preprint arXiv:1807.03748}, 2018.

\bibitem{parisi2019continual}
German~I Parisi, Ronald Kemker, Jose~L Part, Christopher Kanan, and Stefan
  Wermter.
\newblock Continual lifelong learning with neural networks: A review.
\newblock {\em Neural Networks}, 113:54--71, 2019.

\bibitem{pomponi2020pseudo}
Jary Pomponi, Simone Scardapane, and Aurelio Uncini.
\newblock Pseudo-rehearsal for continual learning with normalizing flows.
\newblock {\em arXiv preprint arXiv:2007.02443}, 2020.

\bibitem{rebuffi2017icarl}
Sylvestre-Alvise Rebuffi, Alexander Kolesnikov, Georg Sperl, and Christoph~H
  Lampert.
\newblock icarl: Incremental classifier and representation learning.
\newblock In {\em Proceedings of the IEEE conference on Computer Vision and
  Pattern Recognition}, pages 2001--2010, 2017.

\bibitem{russakovsky2015imagenet}
Olga Russakovsky, Jia Deng, Hao Su, Jonathan Krause, Sanjeev Satheesh, Sean Ma,
  Zhiheng Huang, Andrej Karpathy, Aditya Khosla, Michael Bernstein, et~al.
\newblock Imagenet large scale visual recognition challenge.
\newblock {\em International journal of computer vision}, 115(3):211--252,
  2015.

\bibitem{tian2019contrastive}
Yonglong Tian, Dilip Krishnan, and Phillip Isola.
\newblock Contrastive representation distillation.
\newblock {\em arXiv preprint arXiv:1910.10699}, 2019.

\bibitem{wu2019large}
Yue Wu, Yinpeng Chen, Lijuan Wang, Yuancheng Ye, Zicheng Liu, Yandong Guo, and
  Yun Fu.
\newblock Large scale incremental learning.
\newblock In {\em Proceedings of the IEEE Conference on Computer Vision and
  Pattern Recognition}, pages 374--382, 2019.

\bibitem{wu2018unsupervised}
Zhirong Wu, Yuanjun Xiong, Stella~X Yu, and Dahua Lin.
\newblock Unsupervised feature learning via non-parametric instance
  discrimination.
\newblock In {\em Proceedings of the IEEE Conference on Computer Vision and
  Pattern Recognition}, pages 3733--3742, 2018.

\bibitem{yu2020semantic}
Lu Yu, Bartlomiej Twardowski, Xialei Liu, Luis Herranz, Kai Wang, Yongmei
  Cheng, Shangling Jui, and Joost van~de Weijer.
\newblock Semantic drift compensation for class-incremental learning.
\newblock In {\em Proceedings of the IEEE/CVF Conference on Computer Vision and
  Pattern Recognition}, pages 6982--6991, 2020.

\bibitem{zhang2020class}
Junting Zhang, Jie Zhang, Shalini Ghosh, Dawei Li, Serafettin Tasci, Larry
  Heck, Heming Zhang, and C-C~Jay Kuo.
\newblock Class-incremental learning via deep model consolidation.
\newblock In {\em The IEEE Winter Conference on Applications of Computer
  Vision}, pages 1131--1140, 2020.

\bibitem{zhang2016colorful}
Richard Zhang, Phillip Isola, and Alexei~A Efros.
\newblock Colorful image colorization.
\newblock In {\em European conference on computer vision}, pages 649--666.
  Springer, 2016.

\bibitem{zhao2020maintaining}
Bowen Zhao, Xi Xiao, Guojun Gan, Bin Zhang, and Shu-Tao Xia.
\newblock Maintaining discrimination and fairness in class incremental
  learning.
\newblock In {\em Proceedings of the IEEE/CVF Conference on Computer Vision and
  Pattern Recognition}, pages 13208--13217, 2020.

\bibitem{zhao2020memory}
Hanbin Zhao, Hui Wang, Yongjian Fu, Fei Wu, and Xi Li.
\newblock Memory efficient class-incremental learning for image classification.
\newblock {\em arXiv preprint arXiv:2008.01411}, 2020.

\bibitem{zhao2021memory}
Hanbin Zhao, Hui Wang, Yongjian Fu, Fei Wu, and Xi Li.
\newblock Memory efficient class-incremental learning for image classification.
\newblock {\em IEEE Transactions on Neural Networks and Learning Systems},
  2021.

\end{thebibliography}
}

\end{document}